\documentclass[letterpaper, 10 pt, conference]{ieeeconf}  

\IEEEoverridecommandlockouts                              

\let\labelindent\relax

\usepackage[normalem]{ulem}	                        
\usepackage[table,usenames,dvipsnames]{xcolor}      
\usepackage{extarrows}                              
\usepackage{enumitem}

\usepackage{amsmath,amsthm,amsfonts,amssymb,amscd, dsfont}

\usepackage[linesnumbered,ruled,noend]{algorithm2e}  

\SetCommentSty{mycommfont}

\usepackage[font=small]{caption}

\usepackage{graphics} 
\usepackage{epsfig} 

\usepackage{comment}

\usepackage{subcaption}
\usepackage{tabulary}
\usepackage{array,multirow}
\usepackage[colorlinks]{hyperref}

\DeclareMathOperator*{\argmax}{arg\,max}

\newcommand{\meshheight}{2cm}
\newcommand{\pcdheight}{1.6cm}
\newcommand{\graspheight}{1.7cm}

\newtheorem*{assumption*}{Assumption}

\newtheorem*{remark*}{Remark}
\newtheorem*{problem*}{Problem}

\title{\LARGE \bf
Active End-Effector Pose Selection for Tactile Object Recognition through Monte Carlo Tree Search
}

\author{Mabel M. Zhang, Nikolay Atanasov, and Kostas Daniilidis
\thanks{The authors are with the GRASP Laboratory,
        University of Pennsylvania,
        3330 Walnut Street, Philadelphia, PA 19104, USA.
        \{zmen@seas, atanasov@seas, kostas@cis\}.upenn.edu.
        Grateful for support through the following grants: NSF-DGE-0966142 (IGERT), NSF-IIP-1439681 (I/UCRC), ARL RCTA W911NF-10-2-0016, and a GSK grant.}
}

\begin{document}

\maketitle
\thispagestyle{empty}
\pagestyle{empty}

\begin{abstract}

This paper considers the problem of active object recognition using touch only. The focus is on adaptively selecting a sequence of wrist poses that achieves accurate recognition by enclosure grasps. It seeks to minimize the number of touches and maximize recognition confidence. The actions are formulated as wrist poses relative to each other, making the algorithm independent of 
absolute workspace coordinates. The optimal sequence is approximated by Monte Carlo tree search. We demonstrate results in a physics engine and on a real robot. 
In the physics engine, most object instances were recognized in at most 16 grasps. On a real robot, our method recognized objects in 2--9 grasps and outperformed a greedy baseline.




\end{abstract}

\section{Introduction}


Tactile sensing for object recognition has been an area of research since the 1980s \cite{gaston1984, grimson1984, allen1990}. Major advances have been slow, partly due to the sparse nature of touch, which requires more sensing time for a large area coverage, plus motion planning and physical movement time. Additionally, manipulator hardware is expensive. In comparison, vision-based recognition has seen major improvement because of the rich data, rapid information gathering, and low cost.

However, scenarios exist where vision is unreliable, such as dark, dusty, smoky, or blurry underwater environments, transparent and reflective objects, occluded back sides, and objects in a bag. 
In these cases, tactile sensing is a better main modality.
Furthermore, the ultimate goal of manipulation is to contact the object. Starting with contacts early on provides direct physical exteroception that vision cannot achieve.
In fact, physical action is naturally integrated with perception in animals, who use various active tactile sensing organs \cite{prescott2011}. Humans can recover shapes by touch alone.


While some disadvantages of tactile sensing can be compensated by better hardware, 
others can be compensated by efficient planning and exploitation of the limited input. In fact, Flanagan \textit{et al.} \cite{flanagan2006} found that the key to sophisticated manipulation in the human hand lies more in the accurate prediction of motor commands and outcomes than in rapid sensing. This learning and prediction are the bases of active sensing.
%
Active tactile sensing had early work in tandem with passive vision \cite{allen1985, stansfield1988, allen1988} and alone \cite{schneiter1986}. Active perception, as noted by Bajcsy \cite{bajcsy1992}, involves a selection strategy that trades off between task success and the energy to achieve it.

\begin{figure}[thpb]
  \centering
  \includegraphics[height=4.5cm]{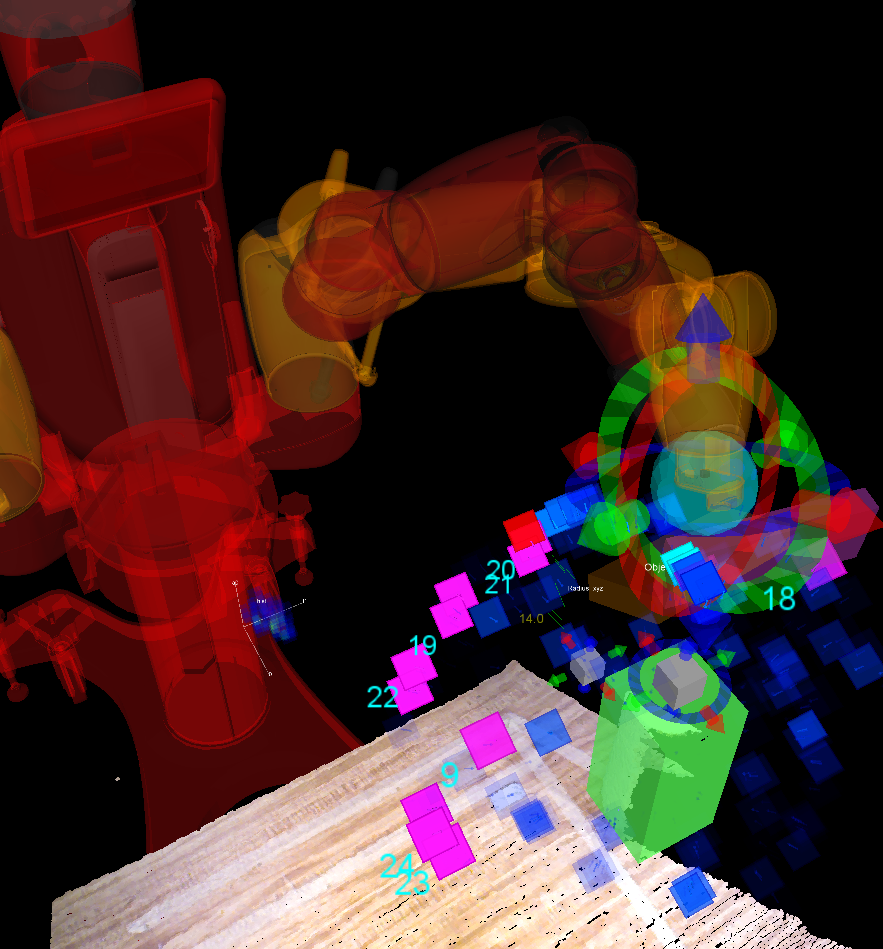}
  \includegraphics[height=4.5cm,trim={5cm 0cm 3.5cm 8cm},clip]{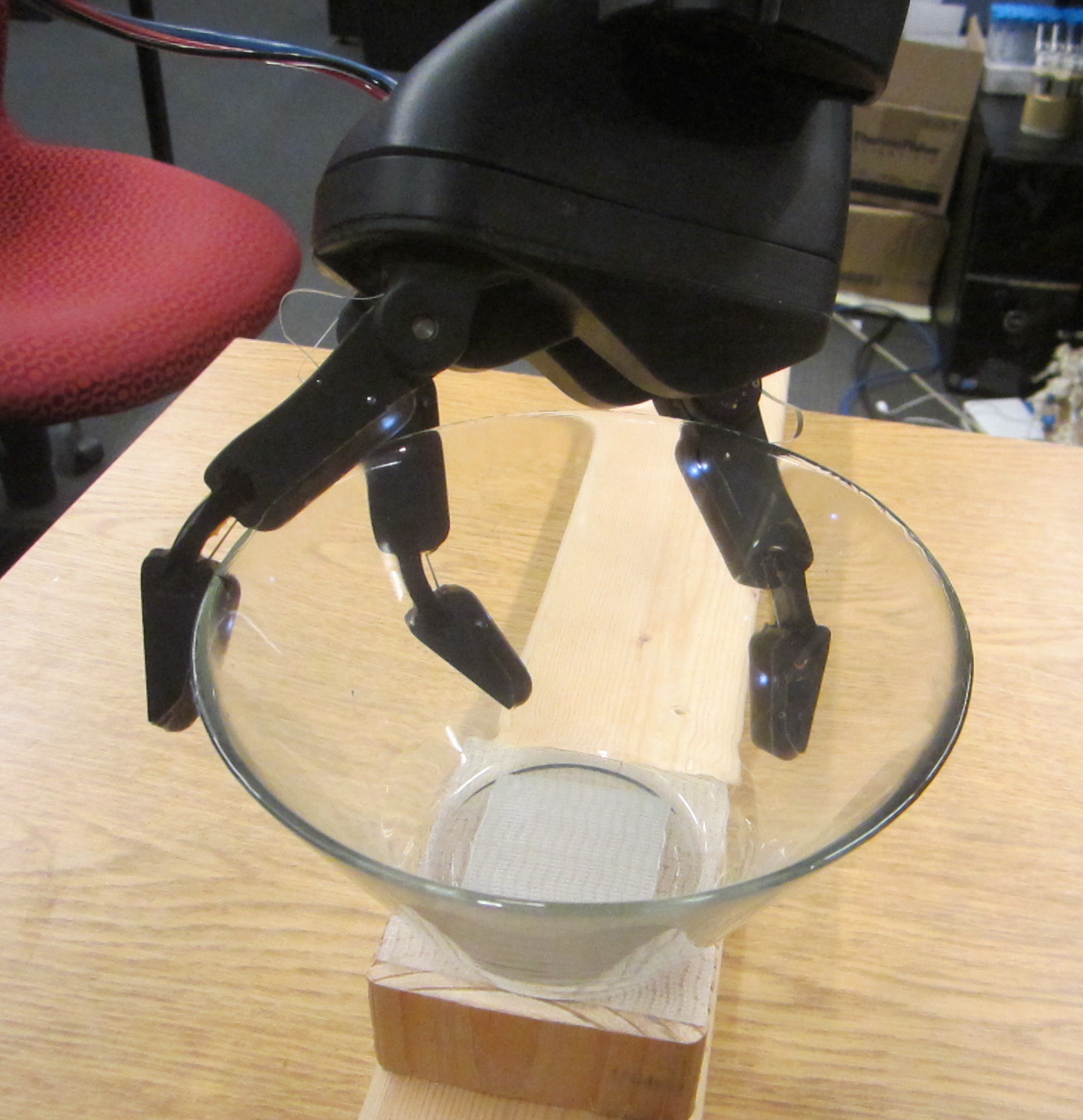}
  \caption{Left: Experiment setup. Right: An adaptively selected pose.}
  \label{fig:intro}
  \vspace{-6mm}
\end{figure}


In this paper, we tackle touch-only object recognition by an autonomous active selection algorithm, which aims to select a minimum number of wrist poses for maximum recognition likelihood.
We formulate the problem as a Markov Decision Process (MDP) and optimize for such a policy.

Our core idea is that consecutive tactile features observed on an object are related to the robot movements in between. Local features are not unique, repeating at symmetric parts and similar curvatures. Discretizing them across objects creates common features that can be modeled as a probability distribution,
which we condition on observations and actions, independent of large state space dimensionality.

We focus on the active prediction problem and not the recognition. For recognition, we use an existing tactile object descriptor \cite{triangles}, the weakness in which is that recognition required hundreds of systematic touches, unrealistic in practice. Our goal is to eliminate that weakness by strategically selecting a small number of touches to observe the key features. We were able to decrease the number by a magnitude.

The novelty of our active approach has three parts. 
First, unlike typical active models, ours is independent of the state space, by conditioning on observations and actions. 
State space-dependent methods have search times proportional to state dimensionality, posing a limit on state definition.
Second, unlike most active perception approaches, ours is not greedy. 
Third, we solve a high-level perception problem as opposed to a low-level sensor-focused haptics one. We target autonomous object-level recognition with cheap sparse pressure sensors, in contrast to most tactile recognition work, which are material-level recognition with expensive dense multi-modal sensors and predefined strokes. Our algorithmic abstraction 
is not limited to special sensors.
%
We show successful prediction 
in a physics engine and on a real robot.

\section{Related Work}
\label{sec:related_work}




Tactile work has been done for reconstruction, localization, pose estimation, and recognition. 
Our work differs from haptics work on material recognition in three major ways. First, we are solving a perception problem at the object level based on high-level geometry abstracted from sensor inputs, not at the material level that directly uses tactile vibrations. Second, we focus on active prediction of the most useful actions, whereas haptic recognition typically uses predefined motions. Third, we execute the actions autonomously.

Similar to our active end-effector pose selection, active viewpoints have been used to select camera poses to gather information, such as in \cite{doumanoglou2016}.
Sensing with vision only is considered active perception. Our work is closer to interactive perception, which physically contacts the environment \cite{bohg2016}.


Early work have explored touch-only object recognition not involving active planning.
Bajcsy \cite{bajcsy1987} compared human haptic exploratory procedures (EPs) observed by Lederman and Klatzky \cite{lederman1987} to robots, and Allen \cite{allen1990} extended them to a tactile robotic hand.
Gaston \cite{gaston1984}, Grimson \cite{grimson1984}, and Siegel \cite{siegel1991} used Interpretation Trees for recognition and pose estimation.


Active touch has been coupled with vision to various extents.
Allen \textit{et al.} \cite{allen1985, allen1988} used vision to guide active touch to invisible regions and explicitly fused the two for shape reconstruction.
Stansfield \cite{stansfield1988} used an initial visual phase for rough object properties and a final haptic phase for detailed surface properties.
Others explored solely using active touch.
Schneiter \cite{schneiter1986} scheduled sensor motions based on \cite{grimson1984} for recognition.
Maekawa \textit{et al.} \cite{maekawa1992} advanced through grid points for reconstruction as contacts were detected.
Hsiao \textit{et al.} \cite{hsiao2007} partitioned the workspace and represented each region as a state in a POMDP for optimal control policy.

Many recent active learning algorithms greedily maximize information gain (IG) \cite{schneider2009, hsiao2010, saal2010, hebert2013, doumanoglou2016} for 
property estimation or task success.
%
Another recent development is adaptive submodularity, a diminishing return property \cite{golovin2010, golovin2011}.
It was shown that entropy can be submodular and was used to greedily maximize IG for near-optimal touches \cite{javdani2013}.


Work most related to ours in active recognition are Pezzementi \textit{et al.} \cite{pezzementi2011_tro} and Hausman \textit{et al.} \cite{hausman2014}. Both use tree search to select actions. However, Pezzementi's tree was for motion planning, with nodes being collision-free configurations. Hausman's tree nodes were entropy, which were minimized to find optimal poses to move the object into camera view. Our tree nodes are tactile observations, and we select end-effector poses to maximize recognition.


Different from greedy policies, a lookahead policy (\textit{e.g.} tree search)
explicitly optimizes cost and gain several steps ahead
\cite{powell}. Its advantage is that it can avoid jumping on immediate high gains that are also extremely costly, and instead favor less costly actions that yield long-term gain.

Solving for lookahead policy directly is impractically costly, as every possible state in each step ahead needs to be considered.
We tackle this in two ways.
First, we use a Monte Carlo optimization method from reinforcement learning literature \cite{kaelblingRL}.
Second, instead of modeling the state space, we formulate a probability dependent only on the observations and actions. It is considerably lower dimensional and generalizes to any object descriptor and robot platform.

Monte Carlo tree search (MCTS)~\cite{browne2012} has become popular for real-time decisions in AI. It is an online alternative to dynamic programming and uses repeated simulations to construct a tree in a best-first order. 
Kocsic and Szepesv\`{a}ri~\cite{uct} showed that tree policy using the UCT (Upper Confidence bounds applied to Trees) guarantees asymptotic optimality. Feldman and Domshlak~\cite{brue} introduced BRUE, a purely exploring MCTS that guarantees exponential cost reduction.
Silver and Veness~\cite{pomcp} extended MCTS to partially-observable models. MCTS has been used for game solving~\cite{mcts_atari} and belief-space planning in robotics~\cite{Hauser_WAFR10,Sukkarieh_ICRA14,Lauri_ICRA14}, but has not been applied to manipulation.


\section{Problem Formulation}
\label{sec:formulation}

Our goal is to adaptively select a minimum sequence of end-effector poses to correctly recognize an object. The input is contact XYZ only, given by enclosure grasps, useful for sensing the volume and the global shape of an object \cite{lederman1987}. 

\subsubsection{Recognition Descriptor}

We focus on optimizing the sequence of poses and use an existing tactile object descriptor \cite{triangles} for recognition.
We cap the sequence at $t=1:T$ poses. At time $t$, a grasp provides $n$ contact points, resulting in $\binom{n}{3}$ triangles $z_t$ \cite{triangles}. Observed triangles $z_{1:t}$ are binned into a 3D histogram $h_t$. The three dimensions represent triangle parameters, \textit{e.g.} two sides and an angle. The histogram is the object descriptor input to a classifier.

\subsubsection{Active Probability}
\label{sec:formulation_core_prob}

In between two consecutive observations $z_t$ and $z_{t+1}$, the end-effector moves by some action $a_{t+1}$, which we model as $a \in SE(3)$, the translation and quaternion from the current wrist pose to a new one. As the hand moves, the previous ending pose becomes the next starting pose, hence removing the need for world frame coordinates for both the hand and the object pose. Let $c_m(a) \in [0,1]$ denote the movement cost incurred by $a$.

To model the recursive chain of $z_t \rightarrow a_{t+1} \rightarrow z_{t+1} \rightarrow \ldots$, we write the probability distribution $p(z_{t+1} | z_t, a_{t+1}, y)$. 
It is in terms of the next observation $z_{t+1}$, conditioned on the current observation $z_t$, the next action $a_{t+1}$ that leads to $z_{t+1}$, and the unknown object class $y$. 

\subsubsection{Training and Test}

During training (Alg.~\ref{alg:train}), two things are learned for each object: its histogram descriptor $h$ and its $p(z_{t+1} | \cdot)$ distribution above. Training is done by moving the robot hand in a grid \cite{triangles} around the object. Actions and observations are recorded to compute the two items.
An action is defined between two wrist poses; $n$ poses yield $n^2$ actions.
Additionally, we train a support vector machine (SVM) classifier (Sec.~\ref{sec:baseline}) on the descriptors. The SVM gives $p(y | h)$, the probability of class $y$ given a histogram.

At test time, the robot chooses its next grasp $a_{t+1}$ (Sec.~\ref{sec:mcts}) based on state $x_t = h_t$. 
Given the current histogram $h_t$, we can obtain the recognition probability $p(y | h_t)$.


\setlength{\intextsep}{2pt}  
\setlength{\textfloatsep}{2pt}  
\setlength{\floatsep}{0pt}  
\begin{algorithm}[htbp]
\caption{Training stage}
\label{alg:train}
\For {each object}
{
  define a grid of wrist poses P wrt object; execute P\;
  store triangle observations $\{z\}$ from contacts\;
  store tallies of observations $\{z\}$ per pose\;
  compute histogram descriptor $h$\;
}
\end{algorithm}

\begin{problem*}[Active Tactile Recognition]
Given an object with unknown class $y$, an initial information state $x_0$, and a planning horizon of $T$ steps, choose a control policy $\pi$ to optimize the cost, which trades off between movement cost and misclassification probability:
\begin{equation} \label{eqn:objective}
\min_\pi C_T(\pi) \triangleq \frac{\lambda}{T} \mathbb{E}\!\left[ \sum_{t=0}^{T-1} c_m(\pi(x_t)) \right]\!\! +\! (1\!-\!\lambda) \mathbb{P}(\hat{y}_T \!\neq y)
\end{equation}
where $\pi$ maps current state $x_t$ to next action $a_{t+1}$, and $\hat{y}_T \!=\! \argmax_y p(y | h_T)$ is the maximum likelihood estimate of the object class.
$\mathbb{P} (\hat{y}_T \!\!\neq\!\! y) \!\!=\!\! 1 \!\!-\! \max_y p(y | h_T)$ is the misclassification probability.
$\lambda \!\!\in\!\! [0,1]$ determines the relative importance
of incurring movement cost (first term) to gather more information vs. making an incorrect recognition (second term).
\end{problem*}


\section{Proposed Approach}
\label{sec:approach}

\subsection{Markov Decision Process (MDP)}
\label{sec:mdp}
The problem can be represented by a finite-horizon MDP defined by $(\mathcal{X},\mathcal{A},\mathcal{T},\mathcal{G}_t)$.
$\mathcal{X}$ is the state space.
$\mathcal{A}$ is a \textit{finite} set of possible actions.
The transition function
\[\mathcal{T}(x_t,a,x_{t+1}) \triangleq \sum_y p(z_{t+1} \mid z_t,a,y) p(y \mid h_t)\]
advances from state $x_t$ to $x_{t+1}$ given action $a$. Histogram $h_t \in x_t$; $z_t$ is determined by $(x_t, a)$; and $h_{t+1} = (h_t, z_{t+1})$ initializes $x_{t+1}$.
\[ 
\mathcal{G}_t(x_t,a,x_{t+1}) \triangleq \begin{cases}
\frac{\lambda}{T} c_m(a), \qquad \qquad \qquad \qquad 0\leq t < T\\
(1-\lambda)(1-\max_y p(y\mid x_t)), \quad \, \, t = T
\end{cases}
\]
is the stage cost. This corresponds to the two terms in Eqn.~\ref{eqn:objective}. 

An MDP can be represented by a graph. Each state is a node, each action is an edge, and 
$\mathcal{T}$ describes how to move from one node to another. 
$\mathcal{G}$ is the cost associated with an edge.
The graph is generated only at test time.

At the start of the process, a random action $a_0$ is selected. This generates observation $z_0$, which initializes histogram $h_0$ at state $x_0$.
Then, we advance through the states by simulating possible actions and following the transitions $\mathcal{T}$ to create nodes and edges (Sec.~\ref{sec:mcts}).
Each node's $z_t$ adds to the histogram.
Over time, the histogram $h_t$ is incrementally filled and resembles the true histogram from training, at which $p(y|h_t)$ would indicate a high probability.

\subsubsection{Relating Observations and Actions}

When we create a new node with $z_t$ at test time, we do not make robot movements to observe an actual $z_t$, because moving after every edge would require hundreds of movements for the entire tree, making the search impractically slow.
Instead, we rely on observations from training.

At the core of our approach is the relationship between observations and actions, modeled by $p(z_{t+1} | z_t, a_{t+1}, y)$, computed from training data (Sec.~\ref{sec:formulation_core_prob}). We trust this relationship to be reliable during training and carries over to test time, at which we directly sample this probability from training.
Note that $p(z_{t+1} | \cdot)$ is independent of the state space $\mathcal{X}$, which is a probability of histograms $p(h_t)$, high-dimensional (1000D) and exponential in search time.
This independence and direct sampling from training 
allows $p(z_{t+1} | \cdot)$ to be computed quickly at test time (Sec.~\ref{sec:running_time}).

\subsection{Monte Carlo Tree Search (MCTS)}
\label{sec:mcts}


This section describes how we 
generate a graph for the MDP at test time and select an optimal policy $\pi$ from it.
We represent the graph by a tree and use a Monte Carlo 
method.
The reader should refer to \cite{browne2012} for an overview of MCTS and \cite{kaelblingRL} for policy search.
The accompanying video animates the concept.
A simple example is shown in Fig.~\ref{fig:tree} and walked through in Sec.~\ref{sec:tree_procedure}.




An optimal policy $\pi$ outputs an optimal action sequence, which is defined as a path with maximum reward (or equivalently, minimum cost) from the root to a leaf. We seek a path that minimizes the objective cost in Eqn.~\ref{eqn:objective}.

After the root is created, Monte Carlo simulations select actions and follow $\mathcal{T}$ to create new edges and nodes.
Each simulation creates one new node.
After a number of simulations, the tree is well populated, and the optimal path is selected.
Each tree depth is a time step, with root at $t=0$ and leaves at $t \leq T$, a defined horizon, or max tree depth.



\subsubsection{Choosing the Next Action $a_{t+1}$}

At time 
$t$, node $x_t$, the next action $a_{t+1}$ is selected as follows.
In an MDP that allows multiple actions per node (known as a multi-arm bandit problem \cite{browne2012}), the choice of an action faces an exploration-exploitation dilemma.
Exploring new or less-seen actions generates unseen parts of the tree, making use of more training data. Re-visiting high-reward actions exploits branches that at the moment seem more likely to be optimal. 

Balancing this dilemma ensures narrowing down the answer while keeping an open mind to see all of the tree.
For contrast, greedy policies 
always exploit the highest-reward action and ignore the exploration half.
%
We use the UCT \cite{uct} upper confidence bound to select actions to balance this dilemma.
At a node $x_t$ at depth $t$, the next action $a_{t+1}$ is:
\begin{align}
a_{t+1} = \arg\max_a \, \left( (1-C_{t_a}) + c \sqrt{ \frac{2 \ln N}{N_a}} \right)
\label{eqn:ucb1}
\end{align}
where the $C_{t_a} \in [0, 1]$ is $C_T(\pi)$ in Eqn.~\ref{eqn:objective}, computed in previous simulations and stored in node $x_t$. It is the cost of an available action edge $a$ at the $x_t$ (see backpropagation in Sec.~\ref{sec:tree_procedure}). $1-C_{t_a}$ is the reward.
$N$ is the number of times the node has been visited, and $N_a$ is the number of times action $a$ has been followed from the node. 

The first term is exploitation; it favors actions with a high existing reward. The second term is exploration; it penalizes actions that have been followed many times. The two terms are balanced by weight $c$, picked by hand. The result is a well-explored bushy tree.
In entirety, the bound selects an action $a_{t+1}$ that minimizes cost $C_{t_a}$. Together with other actions on a path chosen this way from root to leaf, this minimizes the objective cost $C_T$ in Eqn.~\ref{eqn:objective}.

\subsubsection{Inferring the Next Observation} 

Given an action $a_{t+1}$, the next
observation $z_{t+1}$ is sampled from training data: 
\begin{equation}
  \label{eqn:observ}
  z_{t+1} \!\sim\! p(z_{t+1} | z_t, a_{t+1})
  \!=\!\! \sum_y p(z_{t+1} | z_t, a_{t+1}, y) p(y | h_t)
\end{equation}
This reflects the mapping of the MDP transition function $\mathcal{T}(x_t, a_{t+1}) \rightarrow x_{t+1}$. It describes how to move to the next node $x_{t+1}$, given the current node $x_t$ and next action $a_{t+1}$.
The class $y$ is marginalized out, since the true $y$ is unknown.

\subsubsection{Tree Search Procedure}
\label{sec:tree_procedure}

\begin{figure}[thbp]
  \begin{center}
    \includegraphics[width=.8\linewidth]{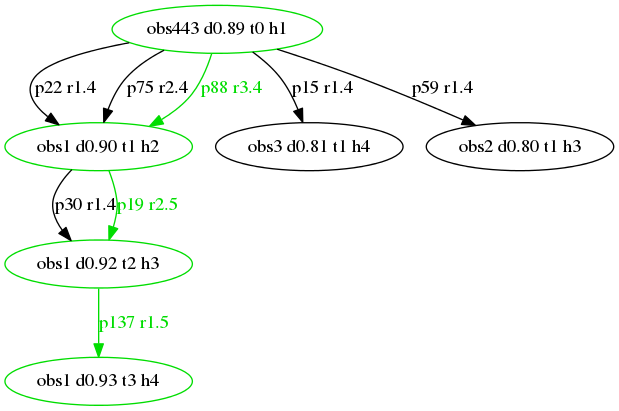}  
  \end{center}
  \caption{A small example tree. Max-reward path highlighted. Node label is observation name, nearest neighbor distance, tree depth $t$, and number of items in histogram $h_t$. NN distance is inversely proportional to $p(y|h_t)$. Edge label is action name $a_{t+1}$ and reward.}
  \label{fig:tree}
\end{figure}

Now we put the pieces together and describe the procedure of each tree search simulation in Algs.~\ref{alg:test},~\ref{alg:tree_search},~\ref{alg:tree_policy}.
Alg.~\ref{alg:test} outlines the top-level test stage procedure.
Algs.~\ref{alg:tree_search} and \ref{alg:tree_policy} outline the tree search and tree policy.
We will walk through a 5-node tree in Fig. \ref{fig:tree}, with horizon $T=3$, the 5 nodes generated from 5 simulations.

Starting with an empty tree, some action is randomly selected and produces obs443 in Fig.~\ref{fig:tree} from training data. This initializes the root at $t=0$, with 1 observation in histogram $h_0$, which happens to be 0.89 distance from the closest object in training.

We will describe one full simulation.
Each simulation starts at the root at depth $t=0$ and must traverse a single path downward until the leaves at horizon depth $t=T$. Each depth contains actions $a_t$ and nodes $x_t$.
The intuition of a path from root to leaf in the real world is a sequence of $T$ actions for the robot to execute.

In each simulation, one new node is created via the choice of $a_{t+1}$ and $z_{t+1}$ (Eqns.~\ref{eqn:ucb1},~\ref{eqn:observ}), outlined in Alg.~\ref{alg:tree_policy} treePolicy.
This means early simulations cannot reach depth $T$ via existing nodes, since the tree is still shallow.
In Alg.~\ref{alg:tree_search}, a recursive function treeSearch traverses the tree, incrementing in depth $t$ (line~\ref{alg:ts_recursive}).
As long as a node exists (line~\ref{alg:ts_exists}), the tree policy is called (line~\ref{alg:ts_tp}) to continue down.
When a desired node does not exist, it is created (line~\ref{alg:ts_node}), which concludes the one node created in the current simulation, and this ends the tree policy. The rollout policy follows (line~\ref{alg:ts_rollout}) and continues to depth $T$ by randomly selecting $a_{t+1}$ at each layer.

In Fig.~\ref{fig:tree}, simulation 1, at $t=0$, action $a_1=p15$ is selected and produces $z_1=obs3$. Since a node with obs3 at depth $t=1$ does not yet exist, it is created, and this ends the tree policy. The rollout policy selects random $a_2$ and $a_3$ that produce temporary nodes $x_2$ and $x_3$, not shown. The rollout policy operates on a temporary subtree that will be discarded at the end of the simulation.
When the rollout policy reaches horizon $T$, the histogram $h_T$ accumulated from observations $z_{0:T}$ on the path we took is fed to the classifier, which outputs the misclassification cost $\mathbb{P}(\hat{y}_T \neq y)$ in Eqn.~\ref{eqn:objective}.

We then trace the path back up to root and backpropagate this cost to store in each node on the path, as follows.
At each depth $t$, reward $1-C_{t_a}$ is updated for action $a$ at node $x_t$.
This is computed by a standard discounted reward, $Q_a = Q_a + (r_a - Q_a) / N_a$, where $Q_a$ is the node's existing reward, and $r_a$ is the raw subtree reward (Alg.~\ref{alg:tree_search} lines~\ref{alg:ts_recursive}--\ref{alg:ts_backpropagate}).

In addition to backpropagation, the objective continues to be computed. At each depth $t$, edge $a_t$ is accumulated to movement cost $c_m(a_t)$ in Eqn.~\ref{eqn:objective}. Intuitively, the reward at node $x_t$ is the sum reward of actions on its current child path.
When we arrive back at the root, the entire objective $C_T$ has now been computed and stored to root under action p15 (Fig.~\ref{fig:tree}). This concludes simulation 1.
In the next 4 simulations, 4 more nodes are created and rewards computed similarly. 
The more simulations, the bushier the tree, and the deeper the branches reach.

After many simulations, the tree search ends by extracting the optimal 
path (Alg.~\ref{alg:test} line \ref{alg:test_selpath}). Starting at the root, simply follow the highest-reward edges downward.
This path defines an action sequence that minimizes $C_T$.
The length of the sequence is $\leq T$, as some branches may not reach $T$, \textit{e.g.} two right branches in Fig.~\ref{fig:tree}.
The optimal sequence is executed on a robot to obtain actual observations for recognition.



\begin{algorithm}[htbp]
\caption{Test stage}
\label{alg:test}
object location given\;
load training probabilities $P_{train}$\;
  superimpose training poses $\{p^o\}$ onto test object\;
move robot to a pose $p_0$ that contacts object\;
close grippers; record observation $z_0$; compute histogram $h_0$\;
\For {each tree}
{
  $node_0$ = initNewTreeRoot $(h_0)$\;
  \For {each simulation}
  {
    treeSearch $(node_0, z_0)$\;
  }
  actions = select max-reward root-to-leaf path\;  \label{alg:test_selpath}
  \For {each action $a_t = a_1 : a_T$ in actions}
  {
    move robot to $a_t$; close grippers\;
    record observation $z_t$; update histogram $h_t$\;
  }
  $z_0 = z_T$; $h_0 = h_T$\;
}
\end{algorithm}

\begin{algorithm}
\caption{Tree search}
\label{alg:tree_search}
\underline{function treeSearch} $(node_t, z_t)$\;
  \eIf {$node_t$ exists in tree}
  { \label{alg:ts_exists}
    $a_{t+1}, z_{t+1} =$ treePolicy$(node_t, z_t)$\; \label{alg:ts_tp}
    subtreeReward = treeSearch$(node_{t+1}, z_{t+1})$\; \label{alg:ts_recursive}
    $r = (1 - c_m (a_{t+1})) +$ subtreeReward $\quad$  \tcp{Eqn.~\ref{eqn:objective}}  \label{alg:ts_reward}
    $node_t$.updateReward $(a_{t+1}, r)$\;  \label{alg:ts_backpropagate}
  }
  {
    create $node_t$\;  \label{alg:ts_node}
    $r =$ rolloutPolicy $(node_t, t) \quad$ \tcp{Eqn.~\ref{eqn:objective} $\mathbb{P}(\hat{y}_T \neq y)$}  \label{alg:ts_rollout}
  }
  return r\;
\end{algorithm}

\begin{algorithm}
\caption{Tree policy}
\label{alg:tree_policy}
\underline{function treePolicy} $(node_t, z_t)$\;
  $a_{t+1} =$ argmax$_a$ UCT$(node_t.C_a, node_t.N_a) $ \tcp{Eqn. \ref{eqn:ucb1}}
  $h_t = node_t.h$\;
  \For {each class $y$}
  {
    $p_y = p(z_{t+1} | z_t, a_{t+1}, y) =$ sampleFrom$P_{train} (z_t, a_{t+1}, y)$\;
  }
  $z_{t+1} = $ marginalizeY$(\{p_y\}) \quad$  \tcp{Eqn.~\ref{eqn:observ}}
  return $a_{t+1}, z_{t+1}$\;
\end{algorithm}

\subsection{Implementation}
\label{sec:setup}


The recognition of test objects is performed by alternating between MCTS and robot action execution. Note that we use the term iteration to refer to one tree search and action execution, \textit{e.g.} two iterations means a tree search, an execution, a second tree search, and a second execution. 


The object pose is assumed known and fixed. 
A first wrist pose is randomly selected from the training data, which store poses with respect to the object center. The observation $z_0$ is computed and initializes the root node of the first tree. 
The tree is generated and produces a sequence of relative wrist poses. This sequence is executed on the robot, with enclosure grasps onto the object at each wrist pose. Actual observations are taken, and a histogram is built and fed to the classifier. This completes one iteration. 
Then, the old tree is discarded, and the latest histogram initializes the root node of a new tree. MCTS is performed again on the new tree.


In order to generalize across objects, triangle observations $z_t$ are discretized to their histogram bin centers. This is required to compute $\mathcal{T}$, which needs probabilities for the conditioned $z_t$ for every object $y$. Otherwise, a triangle from one object might not exist in another to provide this probability. Histogram bin sizes are chosen as in \cite{triangles}.

For the movement cost $c_m(a_t)$, we computed the $L_2$ distance for translation and the angle for rotation, then normalized each to $[0, 1]$ and weighed both equally.
We used $c=1$ to weigh exploration and exploitation equally and $\lambda = 0.5$ to weigh movement and misprediction costs equally.

\begin{figure*}[bthp]
  \begin{center}
  \begin{subfigure}{.69\textwidth}
    \begin{tabular}{c@{\hspace{0.2em}} c@{\hspace{0.2em}} c@{\hspace{0.2em}} c@{\hspace{0.2em}} c@{\hspace{0.2em}} c@{\hspace{0.2em}} c@{}}
      \includegraphics[height=\pcdheight]{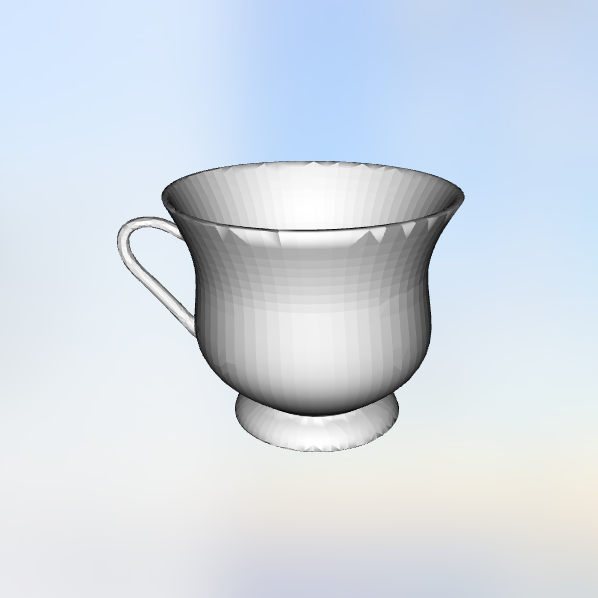} &
      \includegraphics[height=\pcdheight]{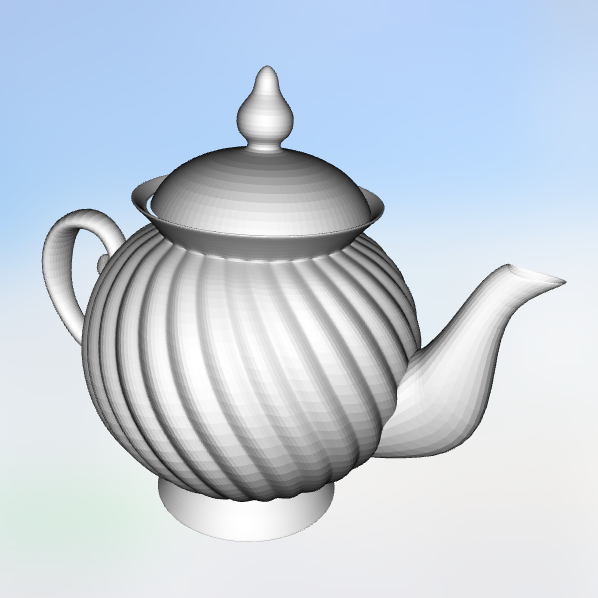} &
      \includegraphics[height=\pcdheight]{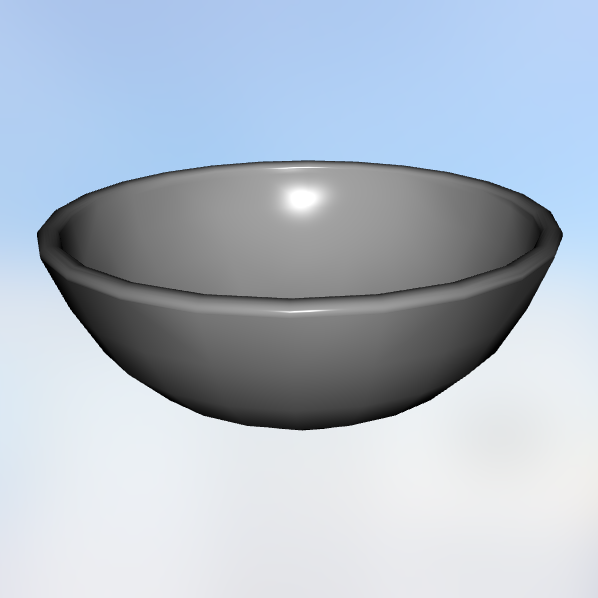} &
      \includegraphics[height=\pcdheight]{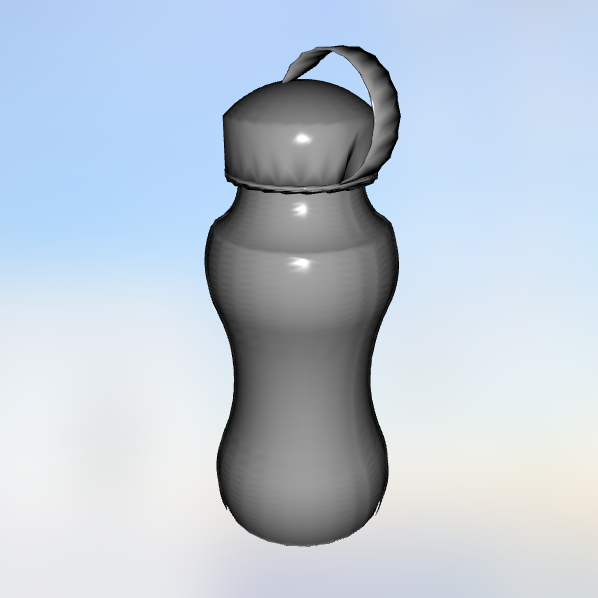} &
      \includegraphics[height=\pcdheight]{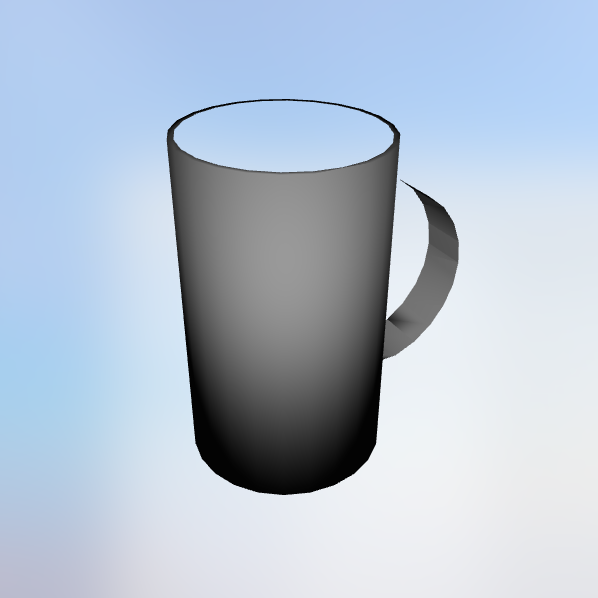} &
      \includegraphics[height=\pcdheight]{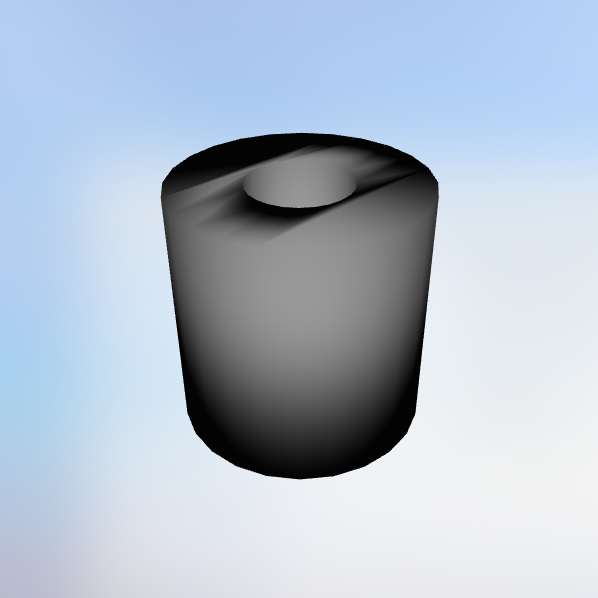} &
      \includegraphics[height=\pcdheight]{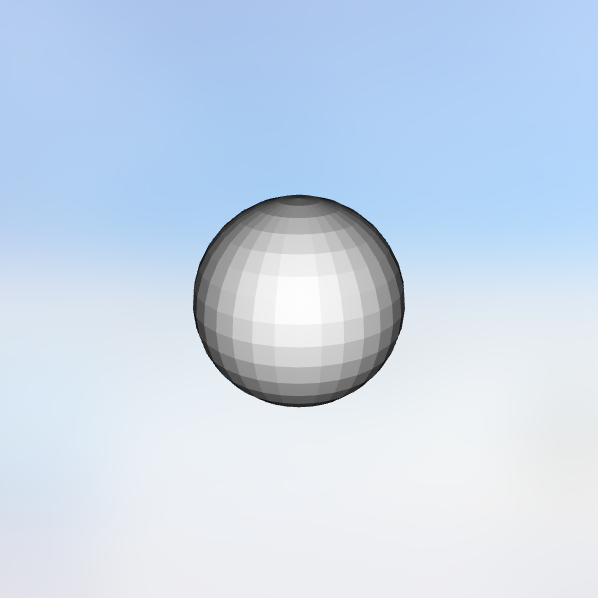}
      \\
      \includegraphics[height=\pcdheight]{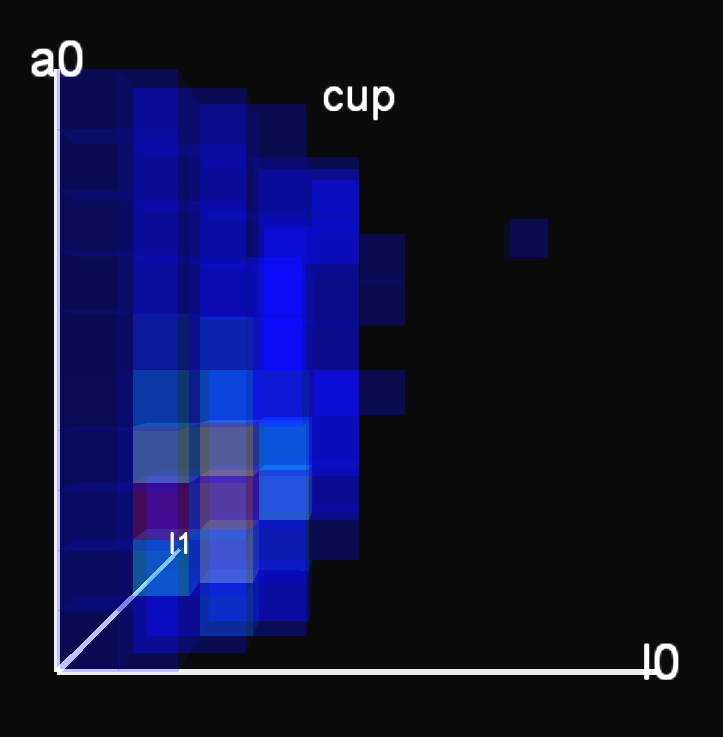} &
      \includegraphics[height=\pcdheight]{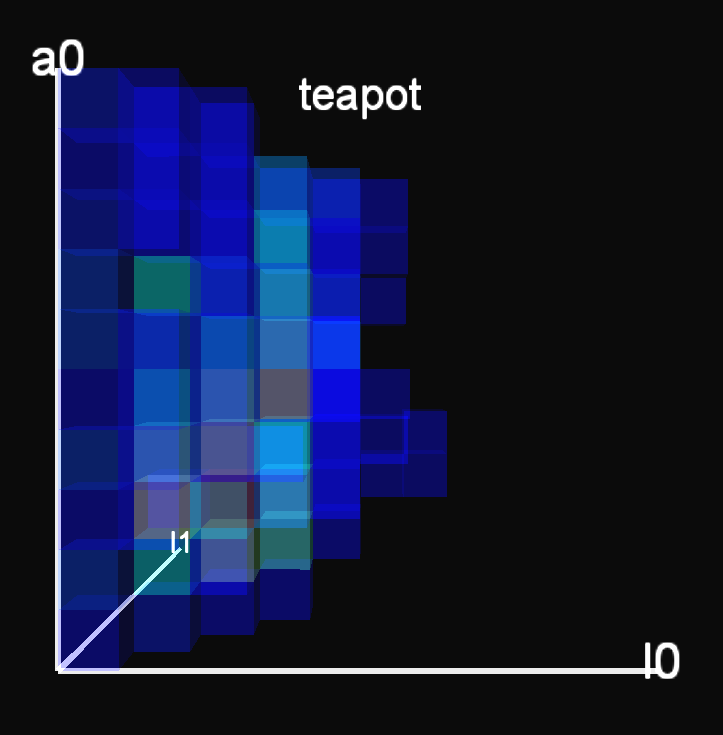} &
      \includegraphics[height=\pcdheight]{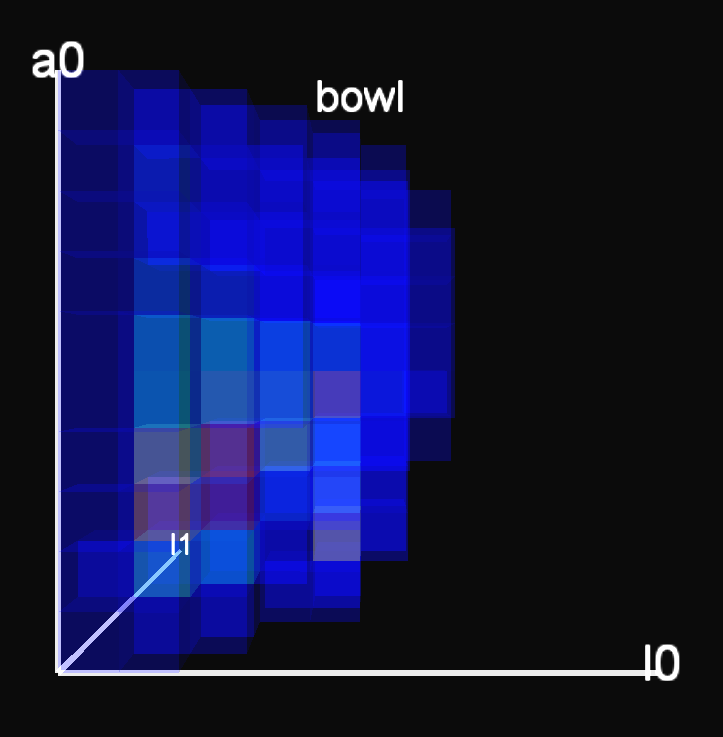} &
      \includegraphics[height=\pcdheight]{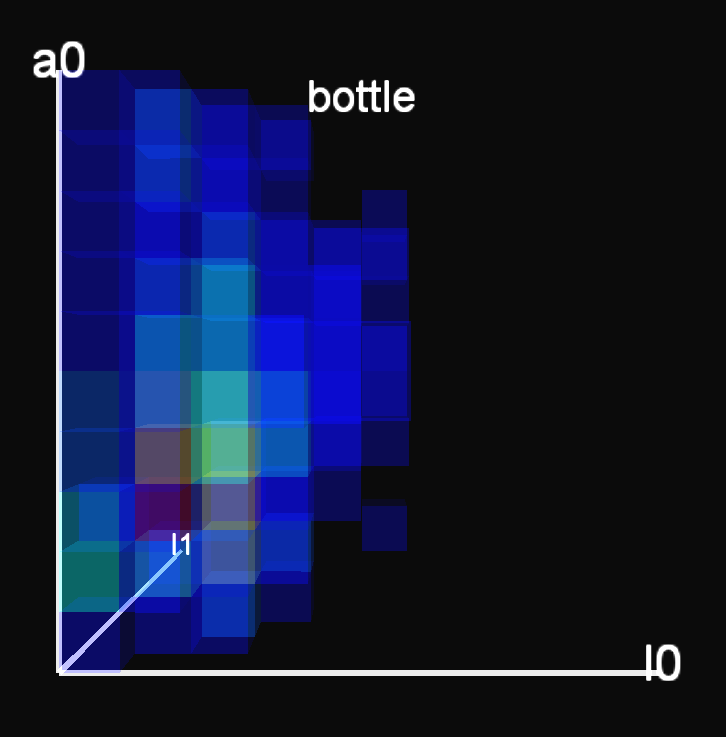} &
      \includegraphics[height=\pcdheight]{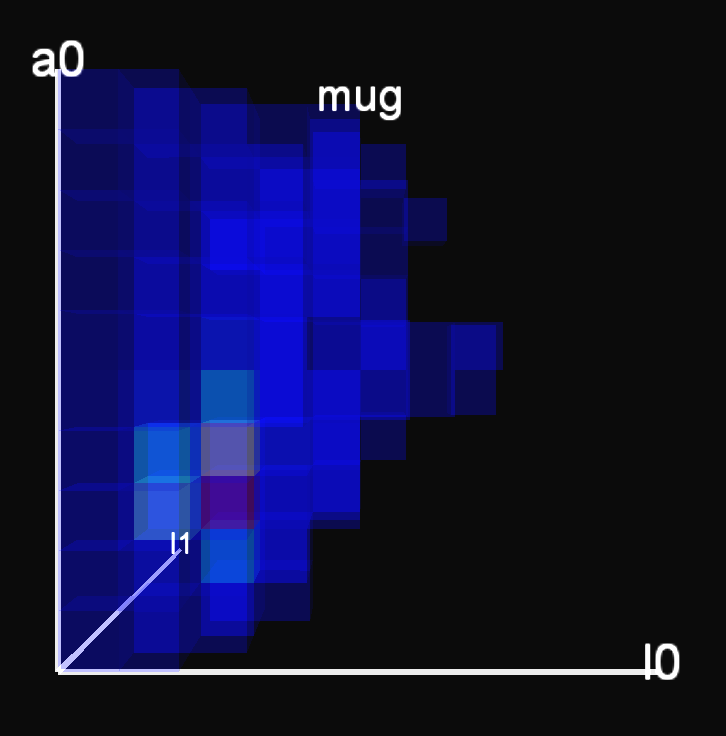} &
      \includegraphics[height=\pcdheight]{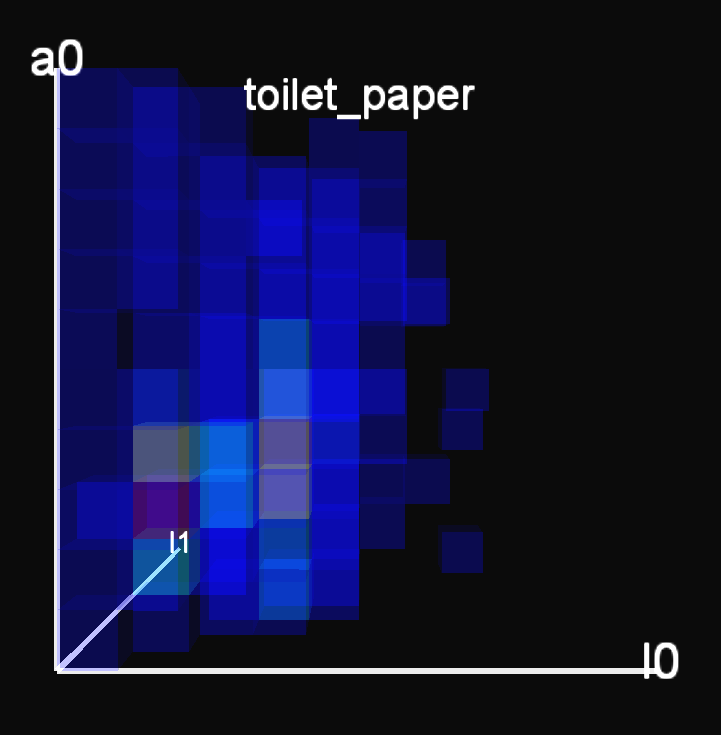} &
      \includegraphics[height=\pcdheight]{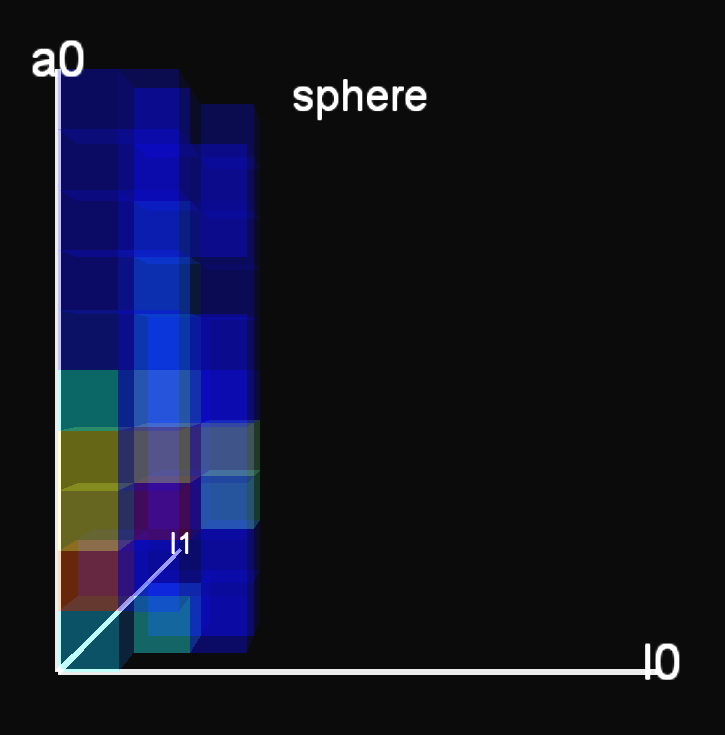}
      \\
      \includegraphics[height=\pcdheight]{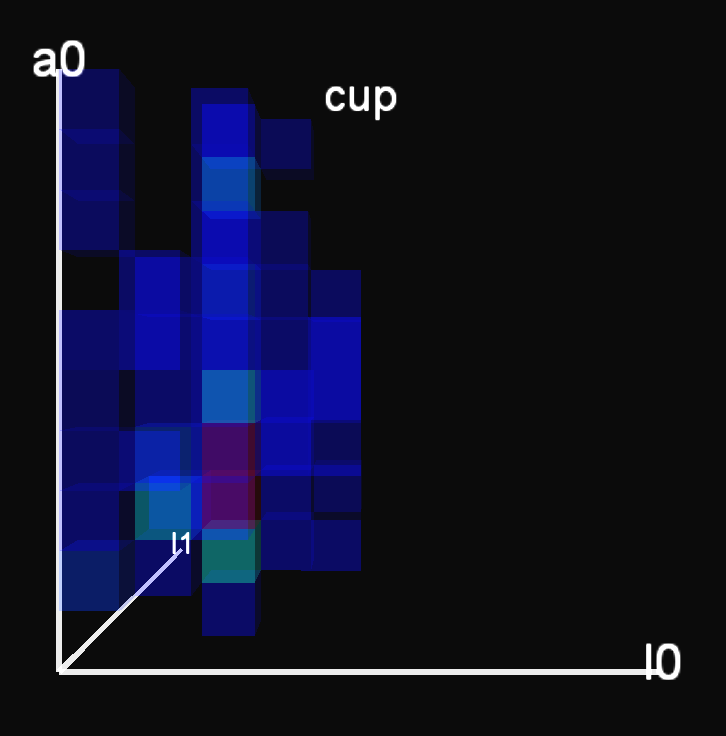} &
      \includegraphics[height=\pcdheight]{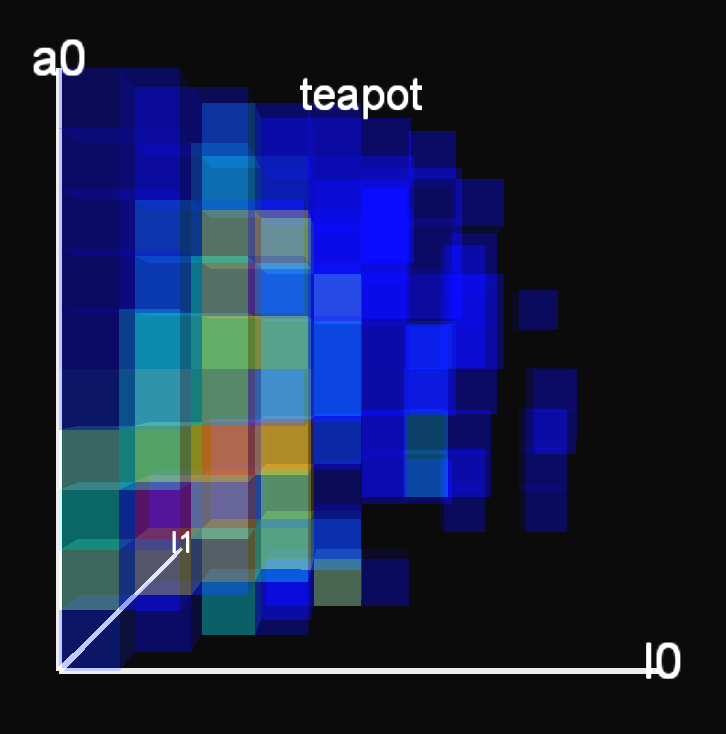} &
      \includegraphics[height=\pcdheight]{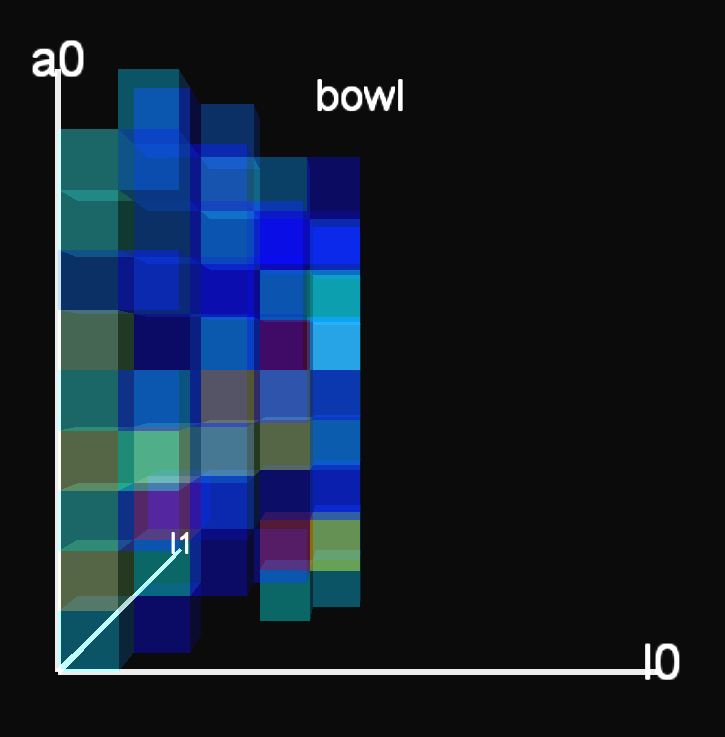} &
      \includegraphics[height=\pcdheight]{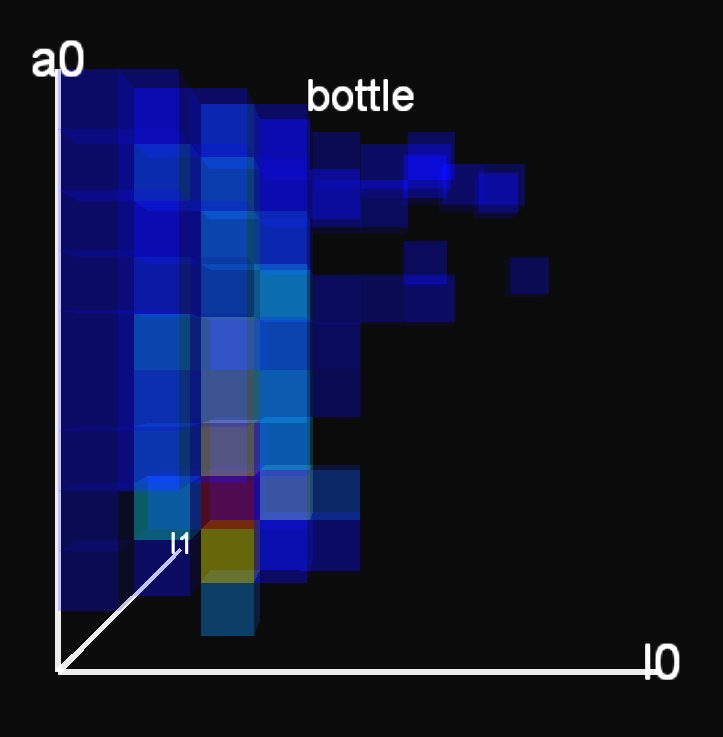} &
      \includegraphics[height=\pcdheight]{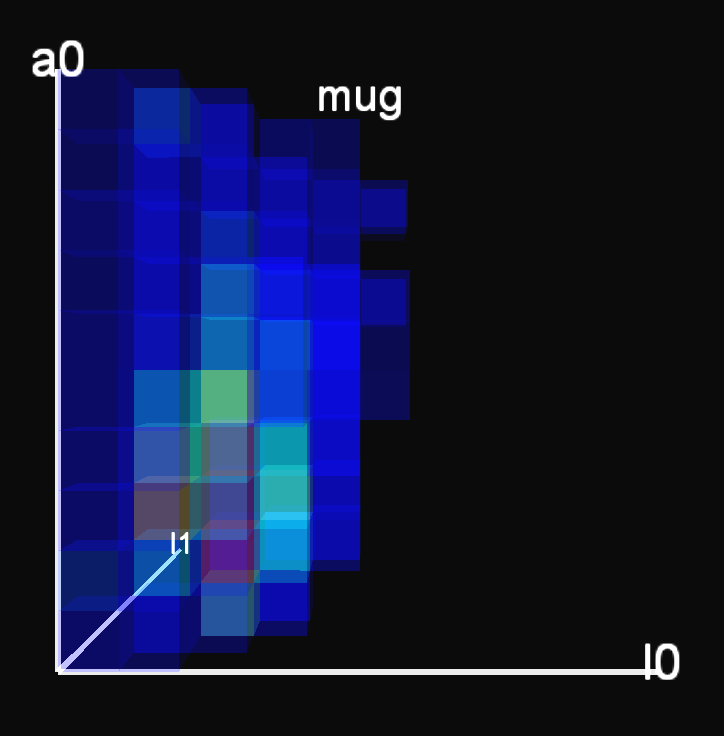} &
      \includegraphics[height=\pcdheight]{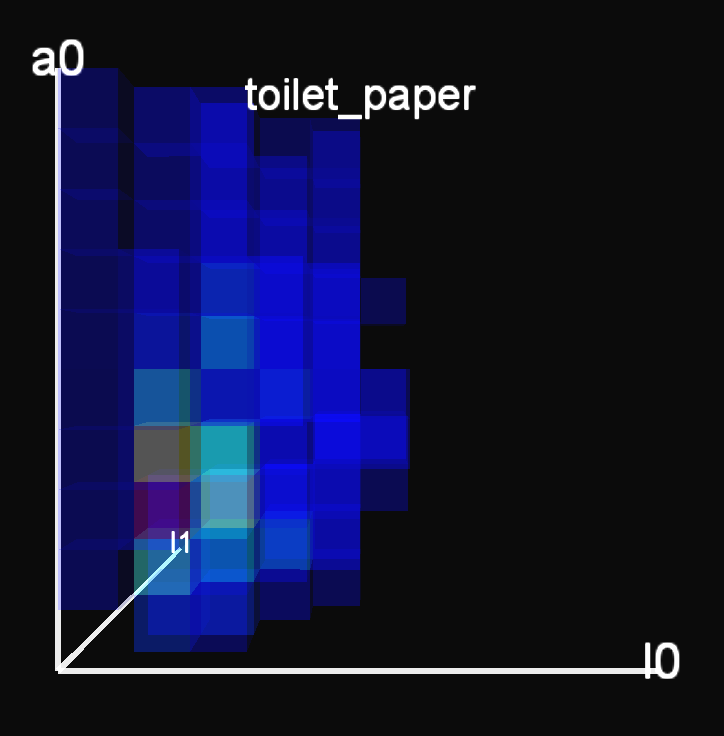} &
      \includegraphics[height=\pcdheight]{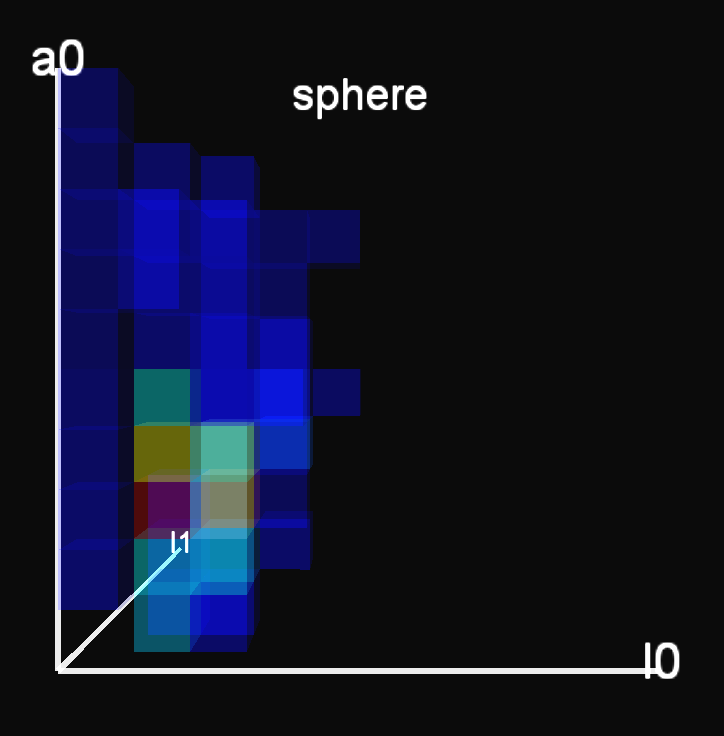}
      \\
      \includegraphics[height=\pcdheight]{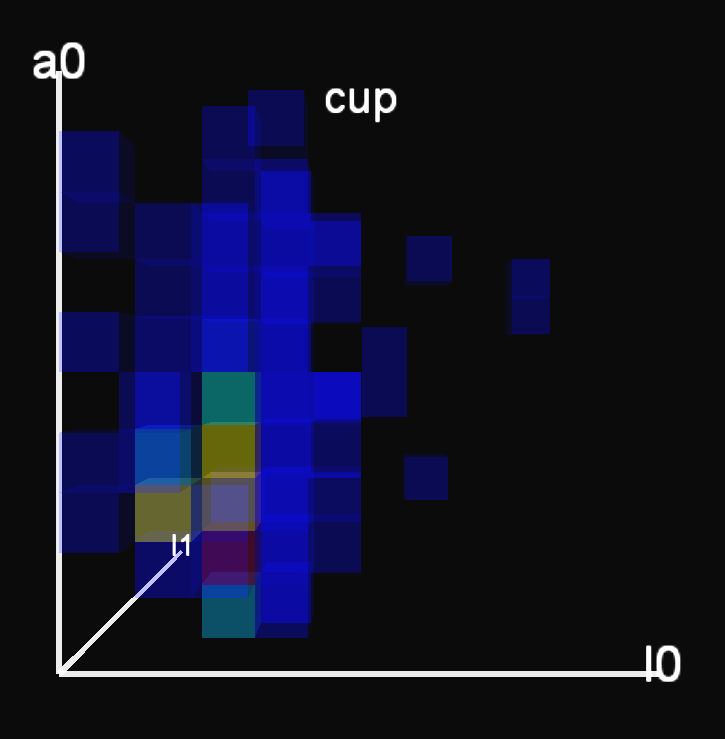} &
      \includegraphics[height=\pcdheight]{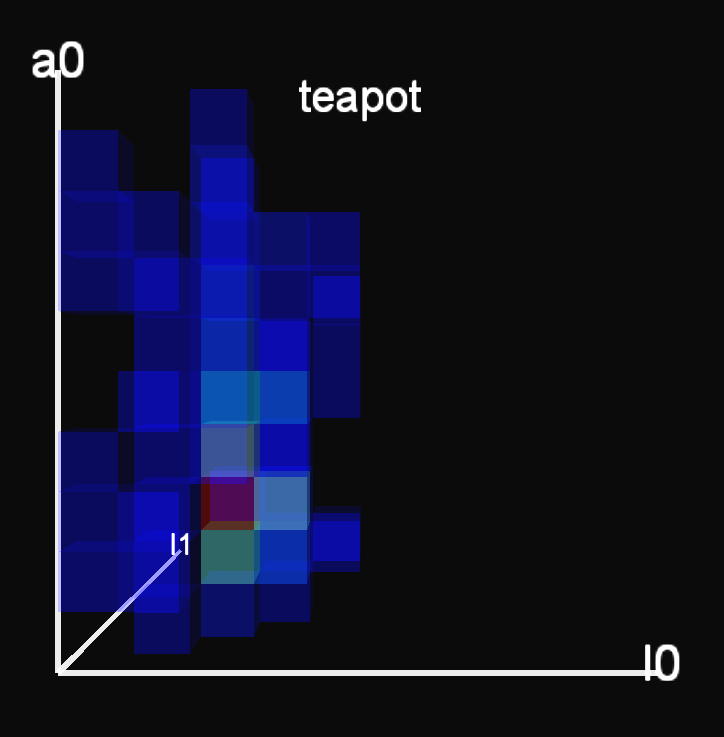} &
      \includegraphics[height=\pcdheight]{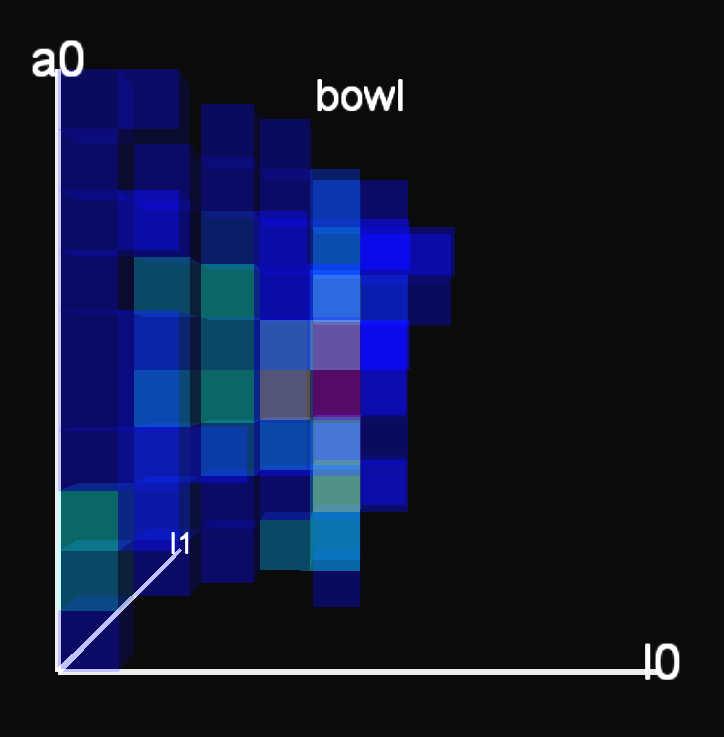} &
      \includegraphics[height=\pcdheight]{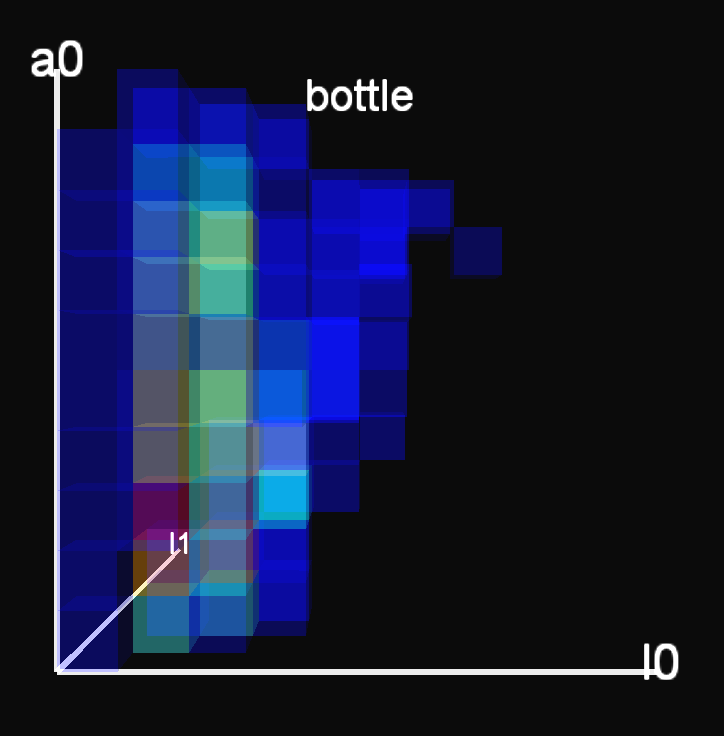} &
      \includegraphics[height=\pcdheight]{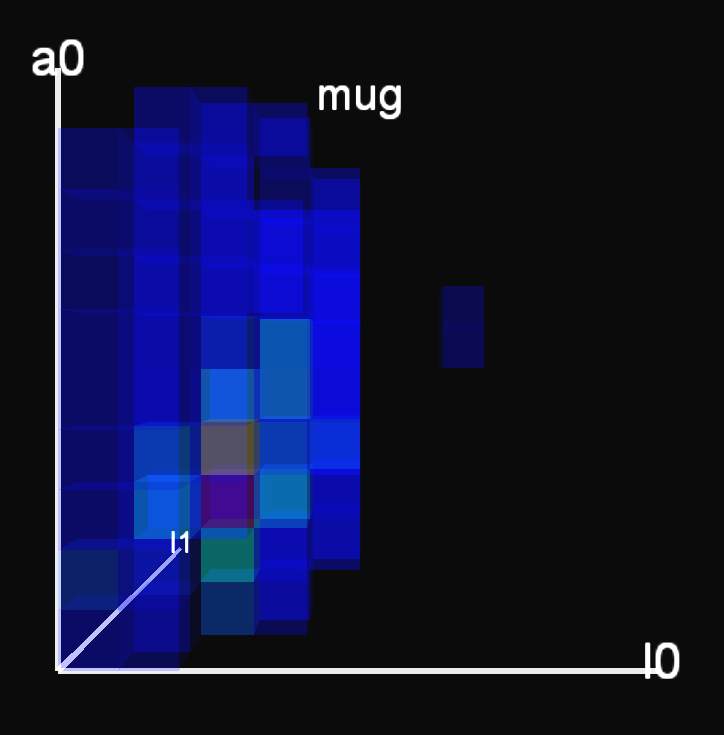} &
      \includegraphics[height=\pcdheight]{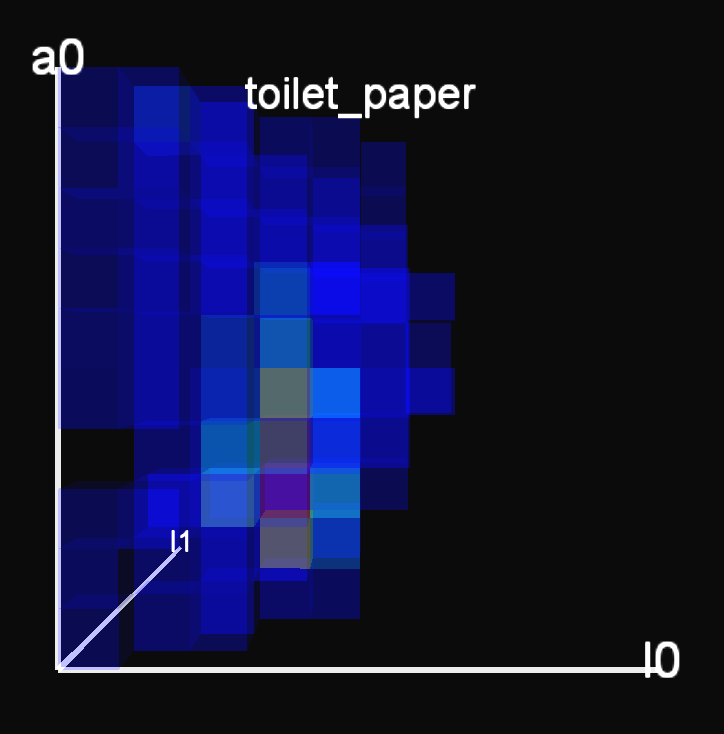} &
      \includegraphics[height=\pcdheight]{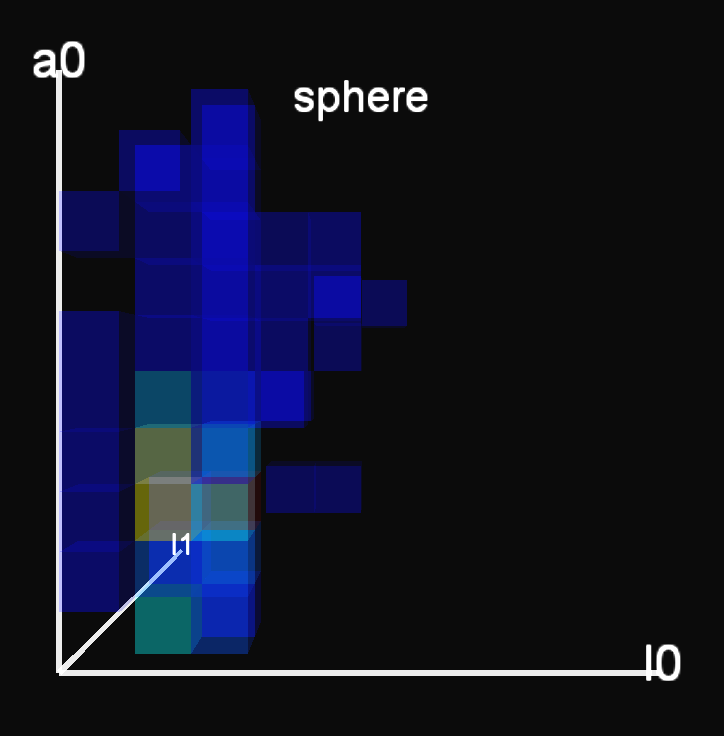}
      \\
    \end{tabular}
    \caption{}
    \label{fig:hists}
  \end{subfigure}
  \vspace{-2mm}
  \begin{subfigure}{.29\textwidth}
    \begin{tabular}{c@{\hspace{0.2em}} c@{\hspace{0.2em}} c@{}}
      \includegraphics[height=\pcdheight]{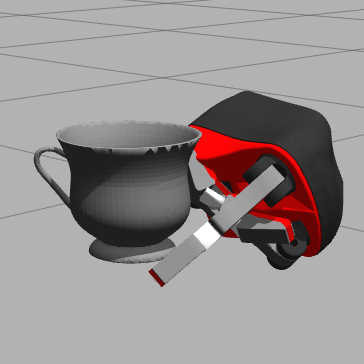} &
      \includegraphics[height=\pcdheight]{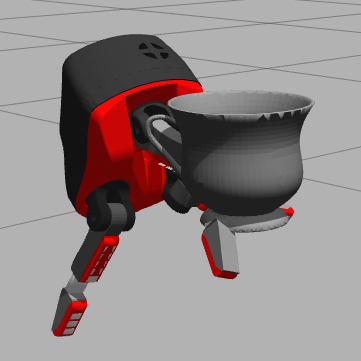} &
      \includegraphics[height=\pcdheight]{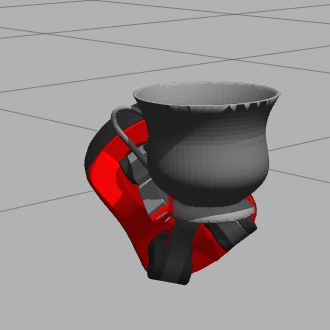}
      \\
      \includegraphics[height=\pcdheight]{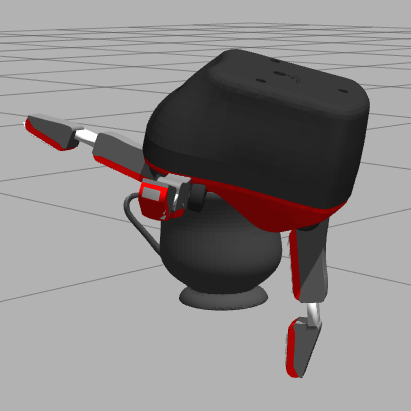} &
      \includegraphics[height=\pcdheight]{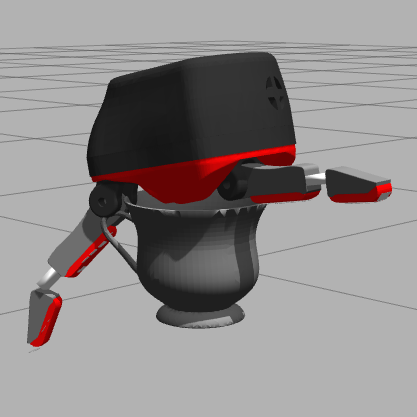} &
      \includegraphics[height=\pcdheight]{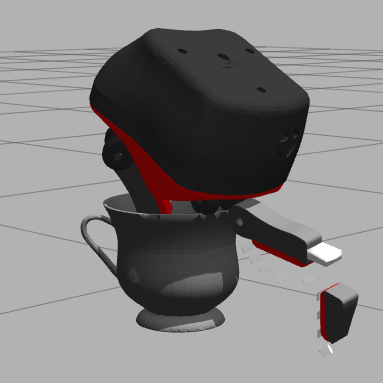}
      \\
      \includegraphics[height=\pcdheight]{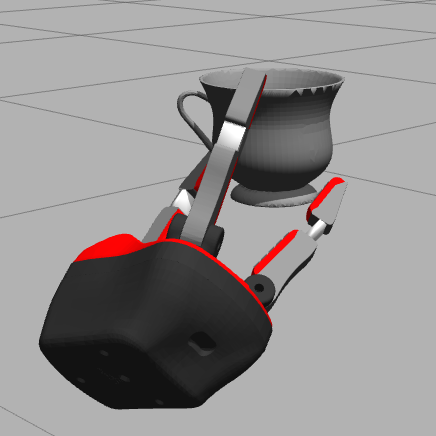} &
      \includegraphics[height=\pcdheight]{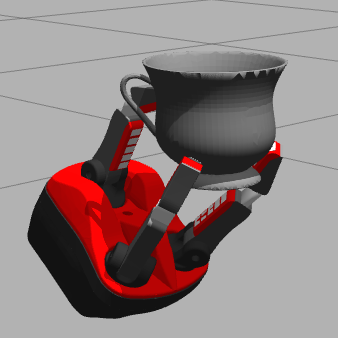} &
      \includegraphics[height=\pcdheight]{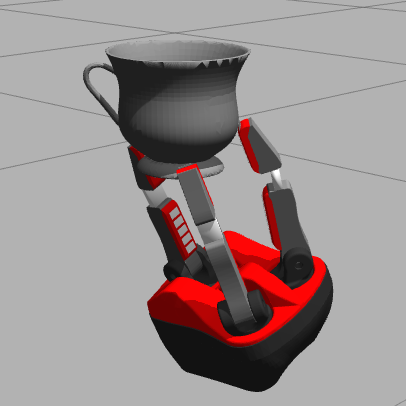}
      \\
      \includegraphics[height=\pcdheight]{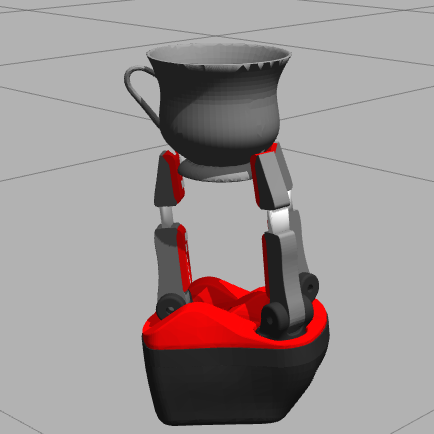} &
      \includegraphics[height=\pcdheight]{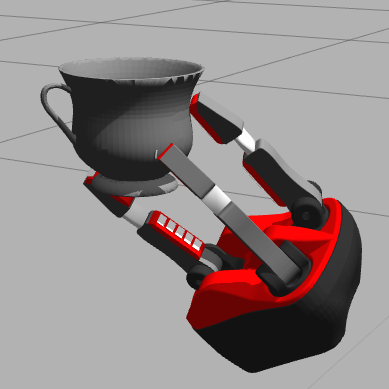} &
      \includegraphics[height=\pcdheight]{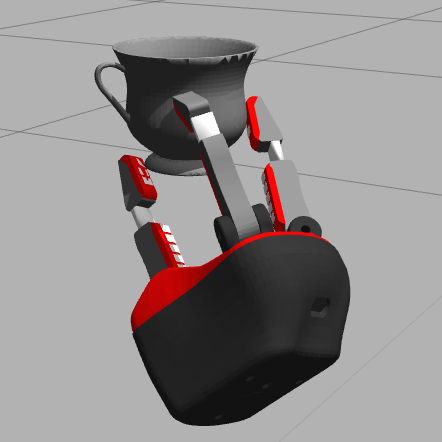}
      \\
    \end{tabular}
    \caption{}
    \label{fig:poses_gz}
  \end{subfigure}
  \vspace{-3mm}
  \end{center}
  \caption{(a). Synthetic meshes and their 3D histograms for visual comparison. From training (row 2), random baseline (row 3), tree policy (row 4). Note shape similarities between cup, teapot, and mug; bottle and mug; mug and toilet paper. Tree policy results resemble true histograms. Best viewed in color. (b). Wrist poses selected by random baseline (top 2 rows) and tree policy (bottom 2 rows) for the cup.}
  \vspace{-5mm}
\end{figure*}

\section{Analysis Using a Physics Engine}
\label{sec:sim}


We validated the active recognition in a physics engine. 
The purpose is to analyze the method itself, without external constraints of a specific robot platform, such as joint and workspace limits. This lets us evaluate the core algorithm using an end-effector with no unreachable poses.

The tactile hardware is a RightHand Robotics ReFlex Beta Hand, which has 9 barometric pressure sensors on each of 3 fingers, 27 total. Each enclosure grasp typically gives non-zero values on 3--6 sensors.
Fig. \ref{fig:hists} shows objects used.

The XYZ positions of the non-zero sensors are used to compute the descriptor for recognition. We used the 3D histogram of triangles \cite{triangles} as descriptor, as mentioned in Sec.~\ref{sec:formulation}. A triangle requires three parameters to be uniquely described. The three side lengths of a triangle are denoted $l_0, l_1, l_2$ from large to small; similarly for the angles $a_0, a_1, a_2$. The following results use the $l_0, l_1, a_0$ parameterization. Because the triangles are a relative measure, the descriptor is independent of object pose and movement.

For physics engine results, we built a stack in Gazebo for the hand's tactile capabilities, including its guarded enclosure, which closes all fingers and stops each finger when a sensor on the finger is in contact. Sensor values were simulated as Boolean contacts. To simulate wrist movement, we teleported the wrist to specified poses. This dramatically reduced the time required, which would otherwise involve motion planning and moving the joint of an arm.

We repeated the MCTS and action execution iterations until rewards in all simulations are depleted to 0, which typically takes 7--9 iterations. When all rewards are depleted, there is only one node in the tree; this happens because we do not allow repeated visits to nominal absolute poses, as they do not provide new observations.

\subsection{Baseline Comparison}

\label{sec:baseline}



To illustrate the need for an active selection, we created a baseline. It uses the same training data as the tree search, except it selects poses to move to at uniform random.
Fig \ref{fig:poses_gz} shows example poses chosen by the baseline and the tree policy for the cup; poses from the latter are more intuitive and capture the global shape better.

\begin{figure*}[thpb]
  \begin{center}
    \begin{subfigure}{.32\textwidth}
      \includegraphics[width=\linewidth]{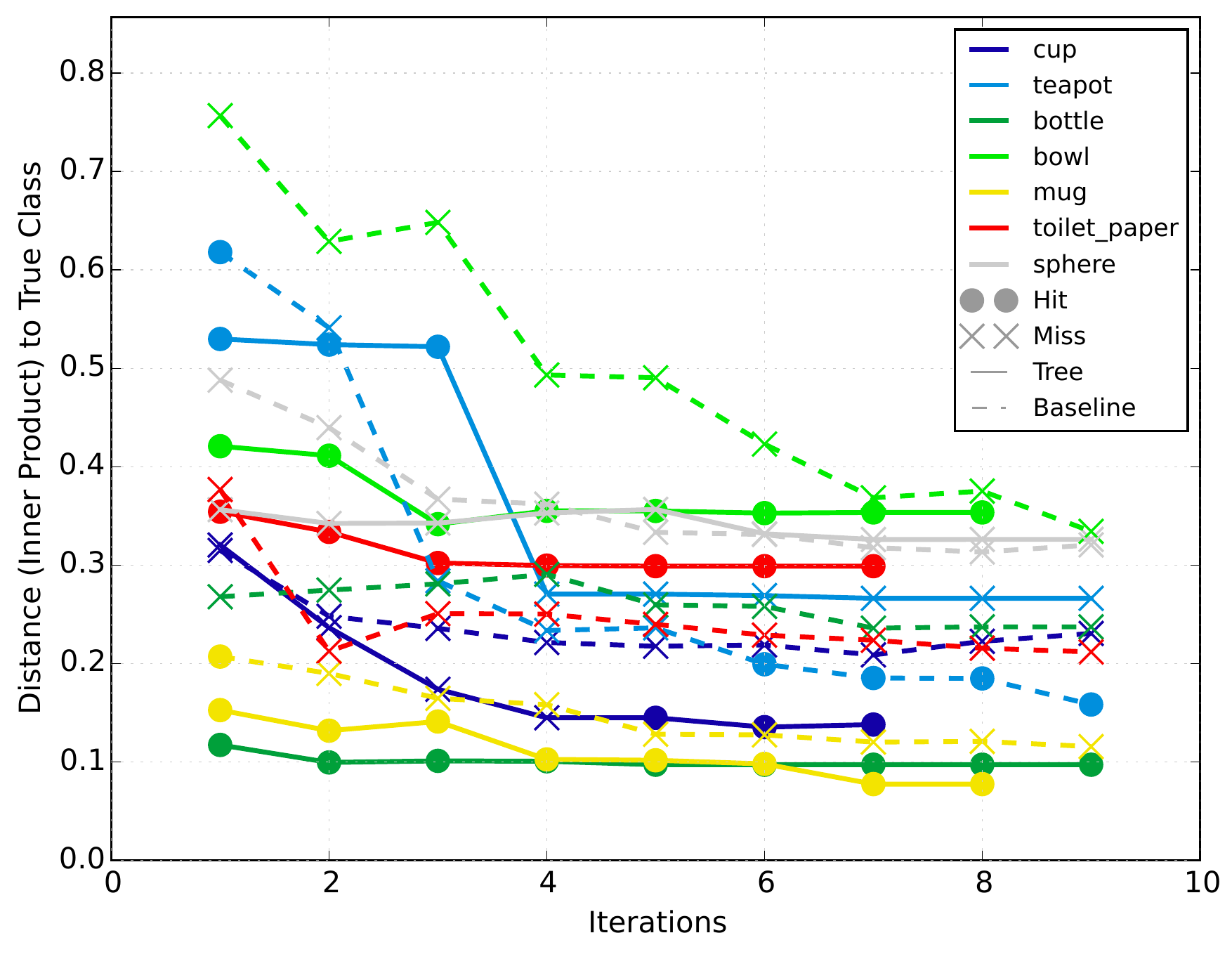}
      \vspace{-7mm}
      \caption{}
      \label{fig:tree_vs_random}
    \end{subfigure}
    \begin{subfigure}{.32\textwidth}
      \includegraphics[width=\linewidth]{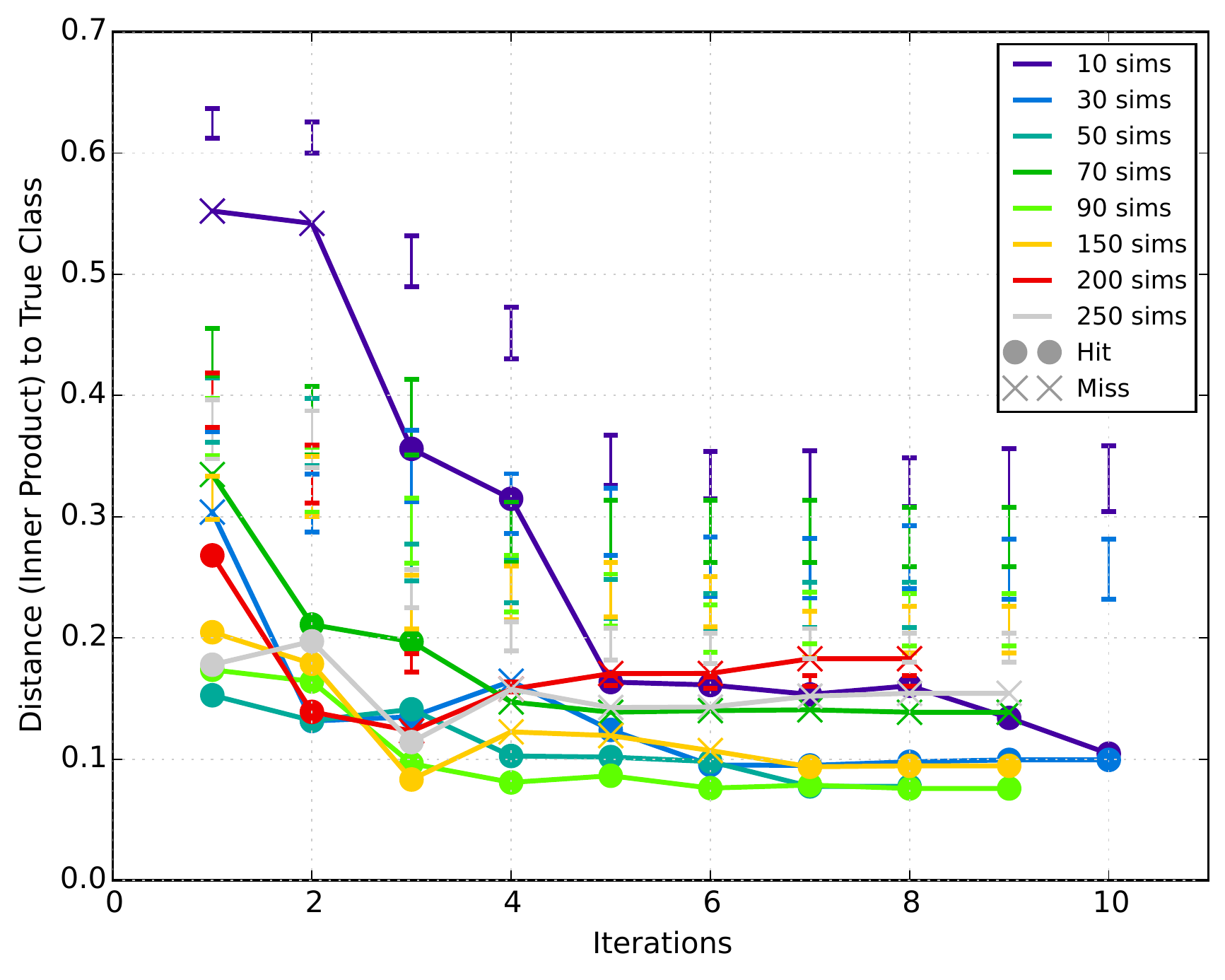}
      \vspace{-7mm}
      \caption{}
      \label{fig:dist_vs_iter}
    \end{subfigure}
    \begin{subfigure}{.32\textwidth}
      \includegraphics[width=\linewidth]{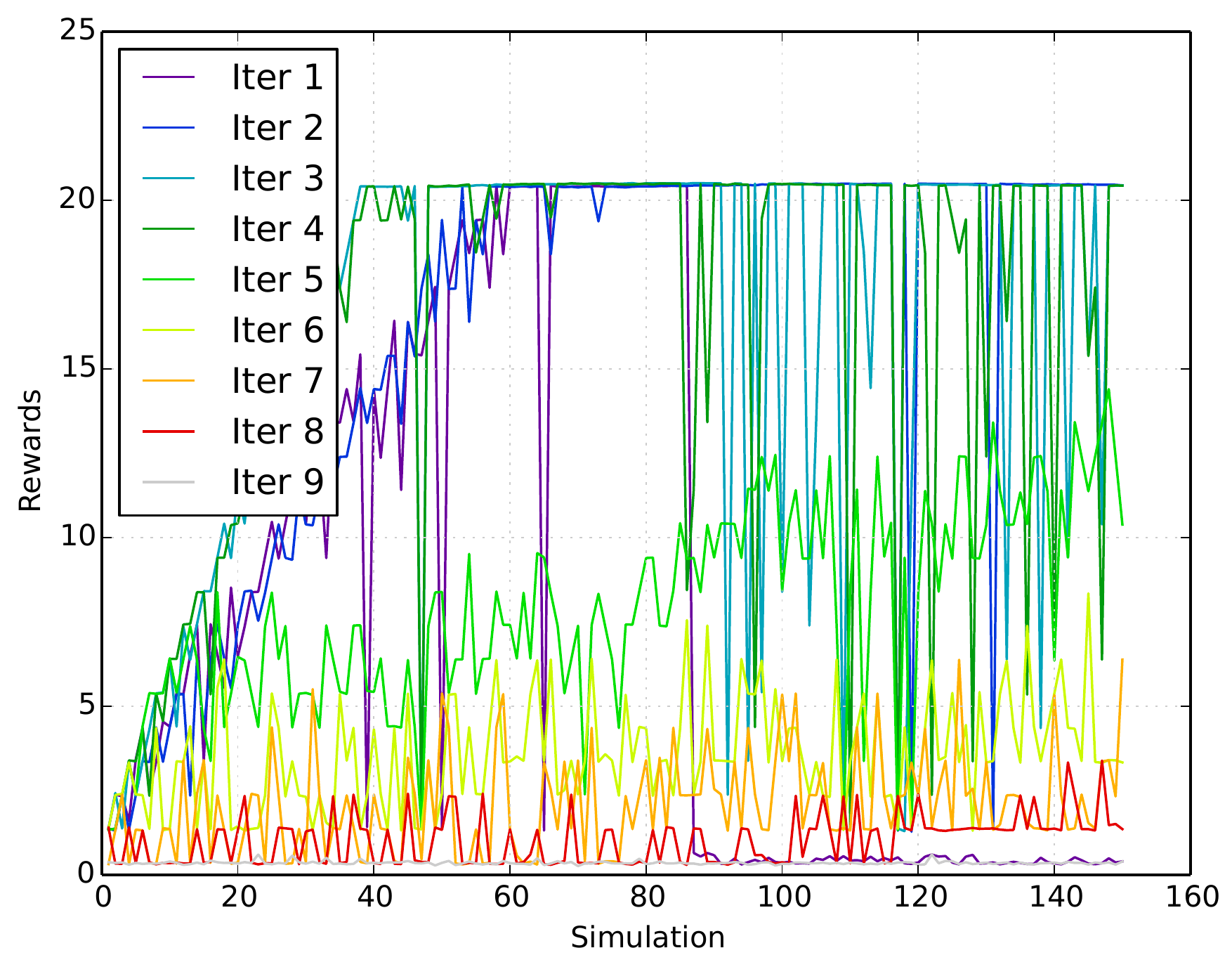}
      \vspace{-7mm}
      \caption{}
      \label{fig:rewards}
    \end{subfigure}
  \end{center}
  \vspace{-7mm}
  \caption{(a). Physics engine results. Distance to true class \textit{vs.} iteration, for baseline (dashed) and tree policy (solid). Dots and crosses show hits and misses. Each curve is an object. Tree policy has more hits and the closest distances. (b). Similar format, for different simulations for mug. Each curve is a simulation setting. Error bars are mean and variance in distances to all objects. (c). Rewards vs. simulations for mug; similar for all objects. Each curve is an iteration. Rewards diminish over iterations as unvisited high-probability poses are exhausted.}
  \vspace{-4mm}
\end{figure*}

Fig.~\ref{fig:hists} and Table~\ref{tab:hist_dist} show the 3D histograms and distances obtained by each method.
Fig.~\ref{fig:tree_vs_random} shows the progression of recognition through the iterations, in the form of distance to true class.
We tried linear SVM, nearest neighbor (NN) with inner product distance, and NN with histogram intersection distance, of which the inner product performed the best.
Tree policy performed better for all objects except teapot, which was correct in iterations 1--3 but diverged to mug in 4--9. This is reasonable as teapot, mug, and cup have similarities in the handle. Sphere was recognized by neither; the cause is evident in Fig.~\ref{fig:hists}, as both were unable to capture the lowest bins in $l_0$.

\begin{figure}[thbp]
  \begin{center}
  \begin{tabular}{c@{\hspace{0.2em}} c@{\hspace{0.2em}} c@{\hspace{0.2em}} c@{\hspace{0.2em}} c@{}}
    \includegraphics[height=\pcdheight]{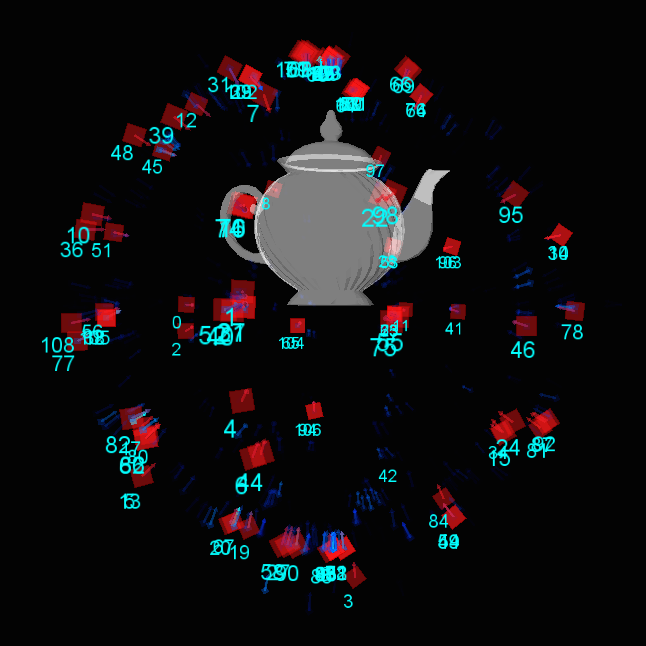} &
    \includegraphics[height=\pcdheight]{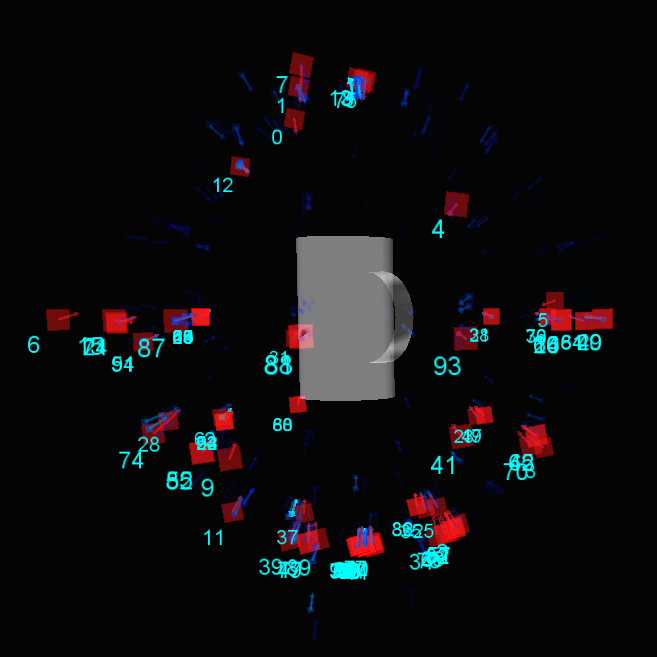} &
    \includegraphics[height=\pcdheight]{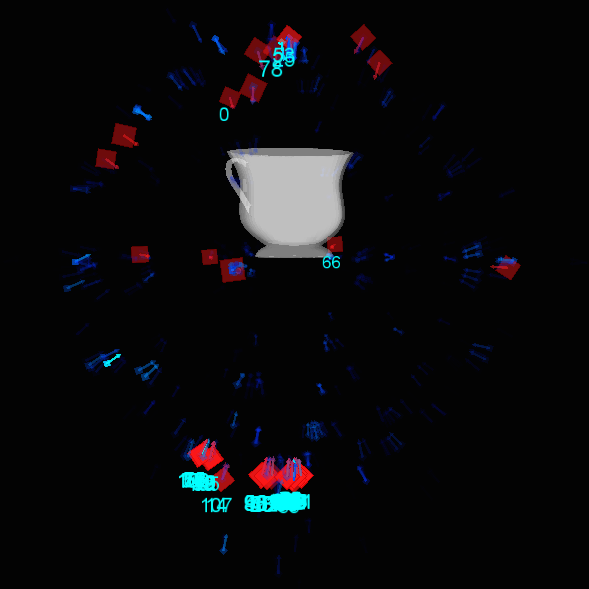} &
    \includegraphics[height=\pcdheight]{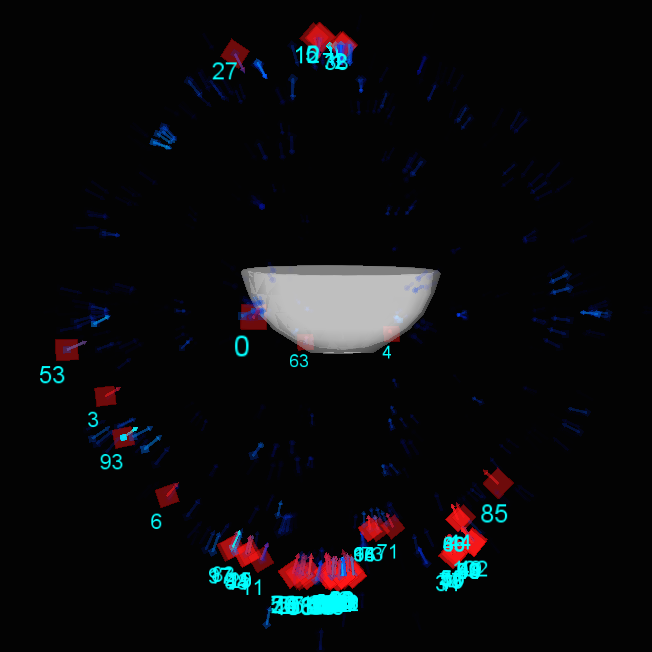} &
    \includegraphics[height=\pcdheight]{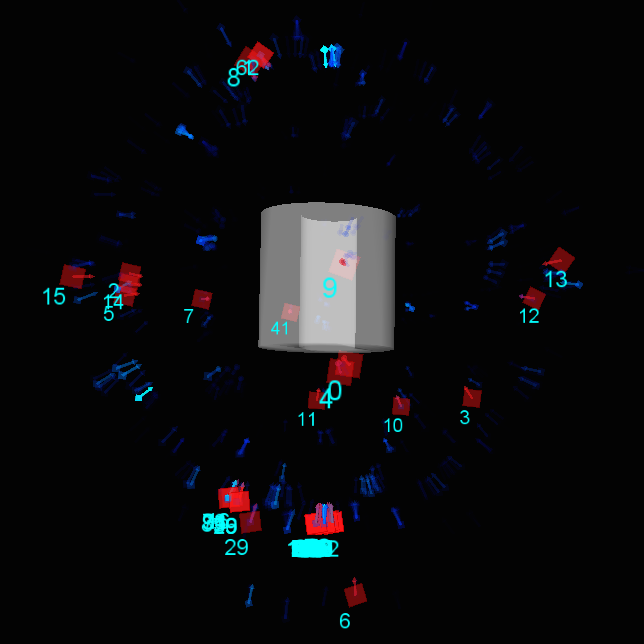}
    \\
    \includegraphics[height=\pcdheight]{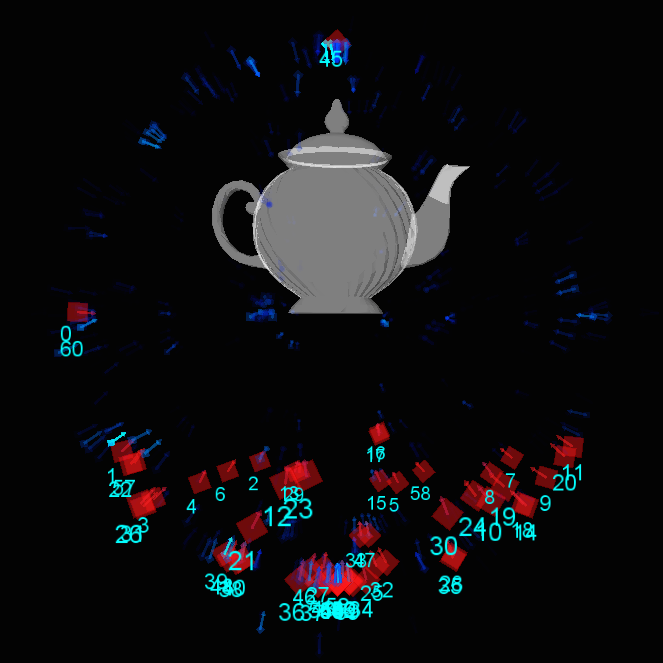} &
    \includegraphics[height=\pcdheight]{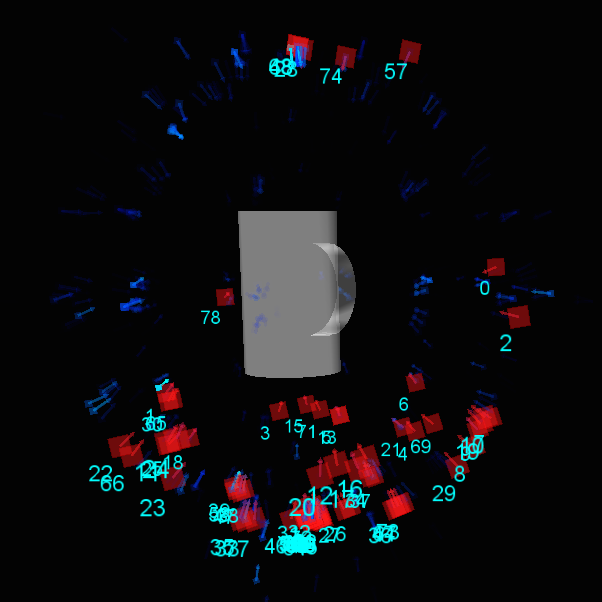} &
    \includegraphics[height=\pcdheight]{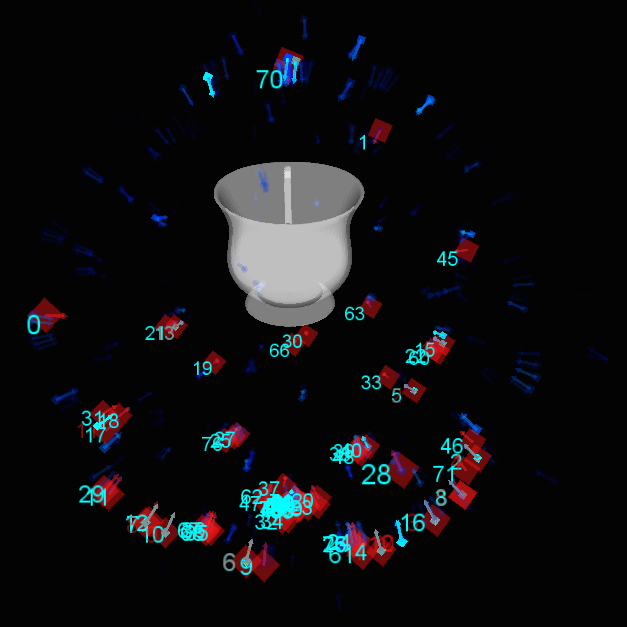} &
    \includegraphics[height=\pcdheight]{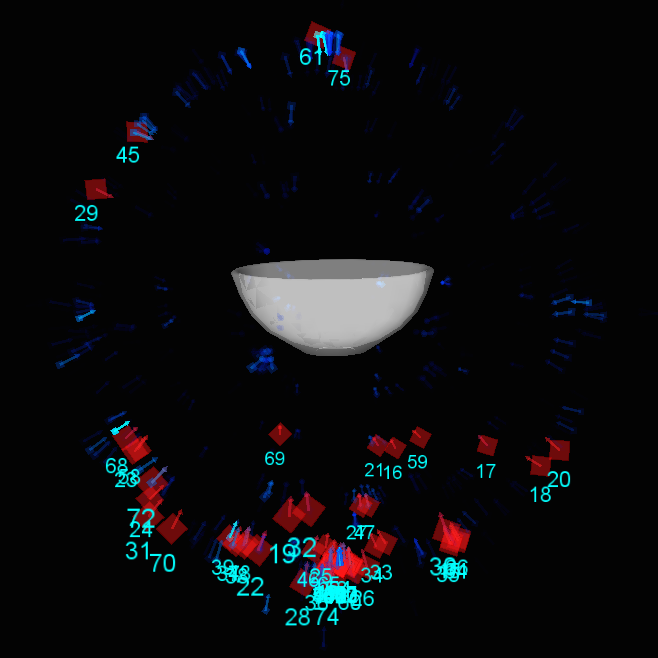} &
    \includegraphics[height=\pcdheight]{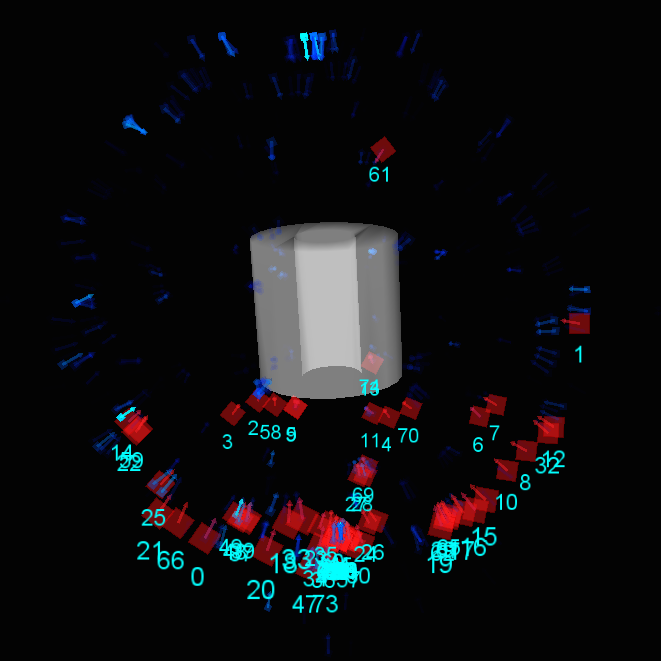}
  \end{tabular}
  \vspace{-4mm}
  \end{center}
  \caption{Poses selected by random baseline (top) and tree (bottom), for teapot, mug, cup, bowl, and toilet paper.}
  \label{fig:poses}
\end{figure}

\begin{figure}[thbp]
  \begin{center}
  \begin{tabular}{c@{\hspace{0.2em}} c@{\hspace{0.2em}} c@{\hspace{0.2em}} c@{\hspace{0.2em}} c@{}}
    \includegraphics[height=\pcdheight]{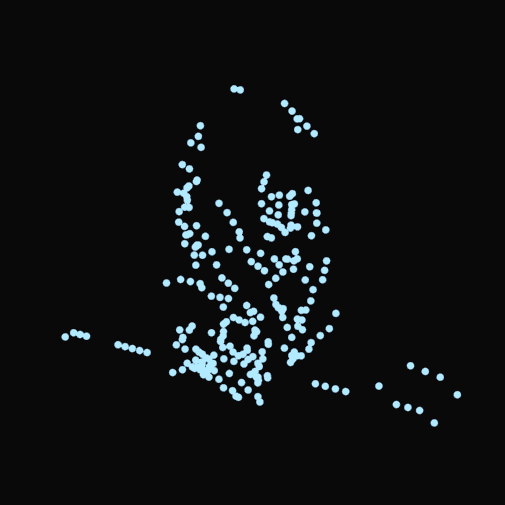} &
    \includegraphics[height=\pcdheight]{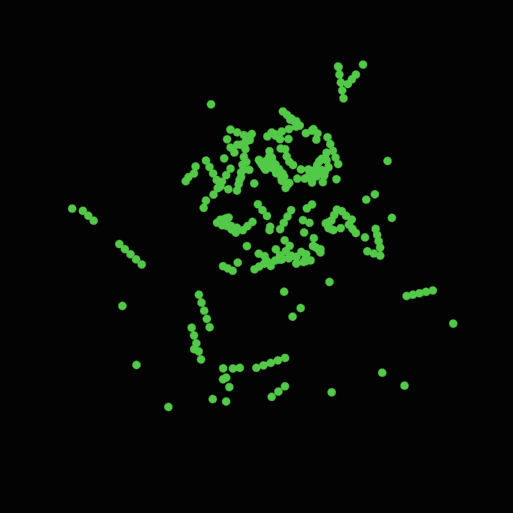} &
    \includegraphics[height=\pcdheight]{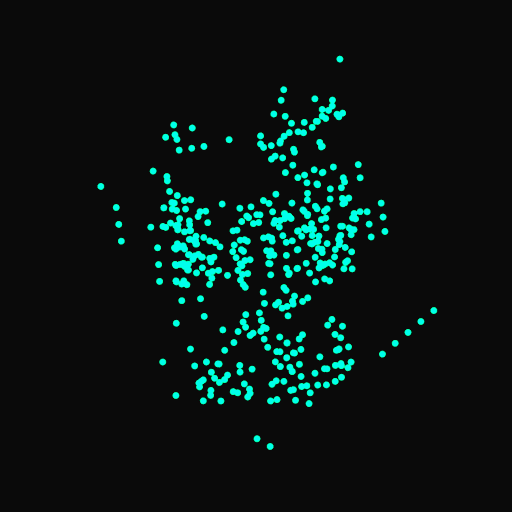} &
    \includegraphics[height=\pcdheight]{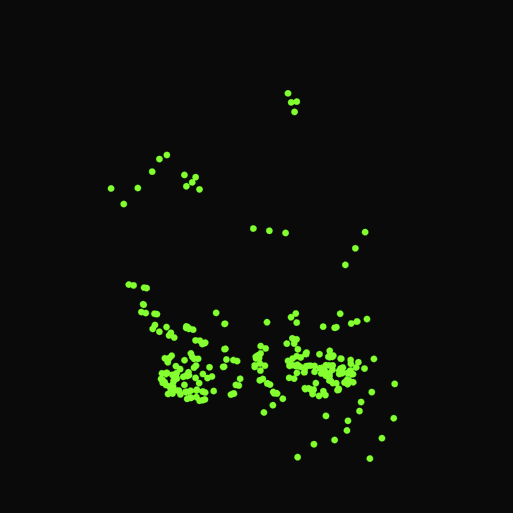} &
    \includegraphics[height=\pcdheight]{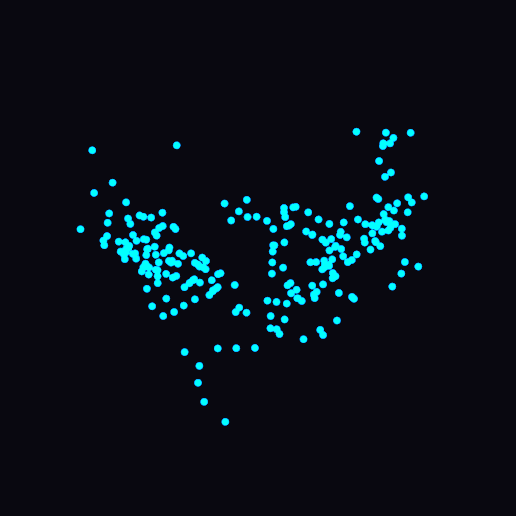}
    \\
    \includegraphics[height=\pcdheight]{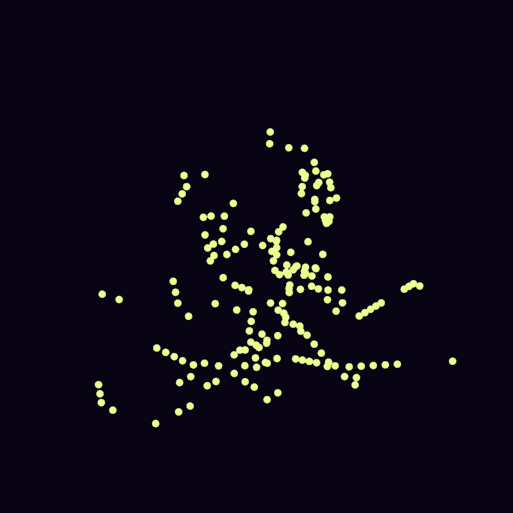} &
    \includegraphics[height=\pcdheight]{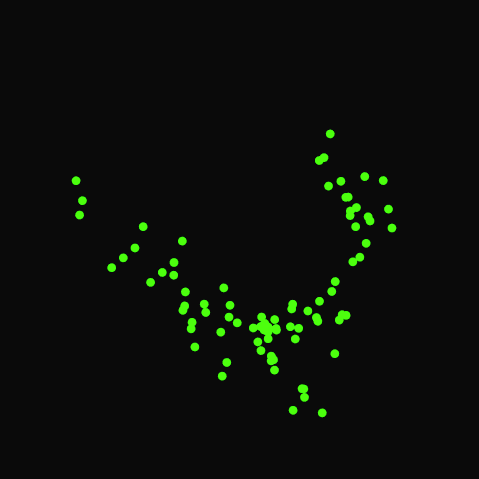} &
    \includegraphics[height=\pcdheight]{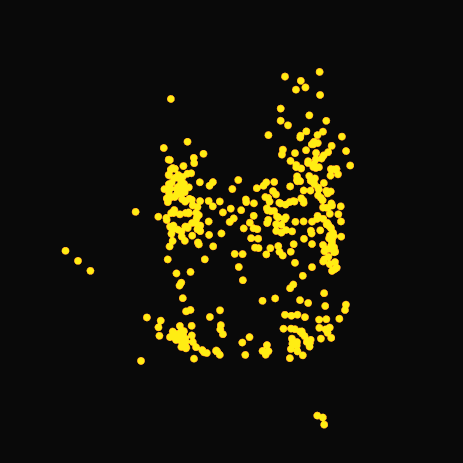} &
    \includegraphics[height=\pcdheight]{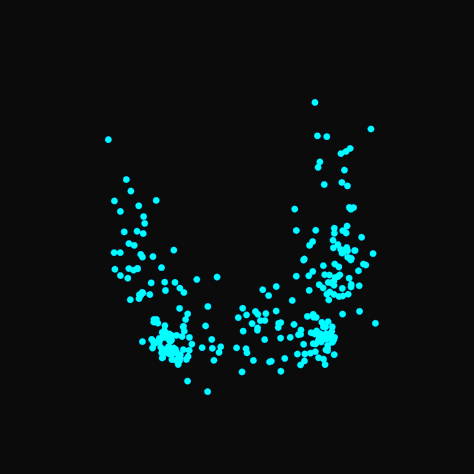} &
    \includegraphics[height=\pcdheight]{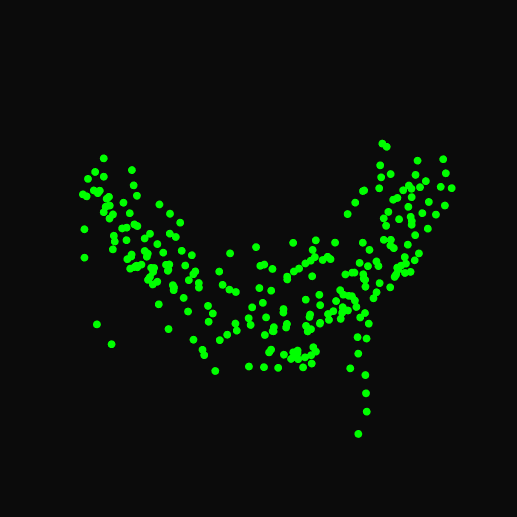}
  \end{tabular}
  \vspace{-4mm}
  \end{center}
  \caption{Actual contacts obtained in physics engine from baseline (top) and tree (bottom), for bottle, teapot, mug, cup, and bowl.}
  \label{fig:pcd}
\end{figure}

Fig. \ref{fig:poses} shows wrist poses selected around the objects, and Fig. \ref{fig:pcd} shows contact points obtained in the physics engine by the two methods. 
Comparing Figs. \ref{fig:tree_vs_random}, \ref{fig:poses}, and \ref{fig:pcd}, even though the baseline sometimes recover a better object appearance, its recognition can still be wrong, \textit{e.g.} bottle, mug.
Tree policy recovered better contact cloud and recognition for cup, bowl, toilet paper.
The only object that the baseline did better in all three was the teapot, most likely because tree policy tends to select poses at the bottom of the object, but the teapot's top half provide identifying information.

Even though the baseline's poses are more evenly distributed, they do not result in better recognition, other than teapot. Fig.~\ref{fig:hists} shows that baseline histograms are more distributed, whereas tree policy's histogram bins are concentrated in the area lit up in the true histograms.
Note that even though some baseline distances in Table~\ref{tab:hist_dist} are closer, \textit{e.g.} cup, toilet paper, its recognition in Fig.~\ref{fig:dist_vs_iter} is incorrect, meaning it is closer to some other class.
This means that the tree policy correctly imposes a bias on the wrist poses selected - poses that result in high recognition certainty, as enforced by $\mathbb{P}(y_T \neq y)$ in the objective. 

\begin{table}[thpb]
  \begin{center}
  \begin{tabular}{c@{\hspace{0.4em}} | c@{\hspace{0.8em}} c@{\hspace{0.8em}} c@{\hspace{0.8em}} c@{\hspace{0.8em}} c@{\hspace{0.8em}} c@{\hspace{0.8em}} c@{\hspace{0.8em}}}
   & cup & teapot & bowl & bottle & mug & tlt ppr & sphere \\ \hline
  Baseline & 0.227 & \textbf{0.160} & 0.375 & 0.237 & 0.121 & \textbf{0.224} & \textbf{0.317} \\
  Tree & \textbf{0.137} & 0.266 & \textbf{0.354} & \textbf{0.097} & \textbf{0.078} & 0.299 & 0.326 \\
  \end{tabular}
  \caption{\label{tab:hist_dist} Test-time histogram distances $\in [0, 1]$ to ground truth histogram from training. Distances are from the last common iteration in Fig.~\ref{fig:tree_vs_random}. Columns correspond to Fig.~\ref{fig:hists}.}
  \end{center}
  \vspace{-1mm}
\end{table}

\subsection{Recognition Performance}

Fig. \ref{fig:dist_vs_iter} shows recognition in the form of inner product distance to true class, for different iterations and simulations for one object. Distances decrease as iterations increase, since the action executions provide increasingly descriptive histograms at the root of new trees.

The error bars show means and variances in distances to all objects. The difference between the error bars and distances to true class indicate that recognition converges early on, in as few as 2 iterations. That is 10--31 poses, average 25 poses across all simulation settings shown. The number of available actions at each node is 1063, pooled across all 7 objects' trained poses. This shows that the tree successfully finds the few essential poses to fill the most discriminative features in the descriptor.

As simulations per tree increase, distances do not necessarily decrease, nor does recognition accuracy increase. 30--90 simulations perform better than 150--250 simulations. This can be due to our restriction of visiting a pose only once. As the available poses are exhausted, the tree policy cannot find an edge at some nodes and has to switch to rollout policy, by design.
This is further reflected in the rewards plot in Fig \ref{fig:rewards}. Before 90 simulations, rewards steadily increase as the tree is more explored; after 90 simulations, deep troughs appear. The troughs are probably due to exhaustion of unvisited poses with high probability on a tree path. Similarly, rewards diminish as iterations increase, when fewer poses are available, 
eventually collapsing to a one-node tree in the last iteration where reward is 0.

Table \ref{tab:nn} shows per-iteration predictions and distances for the cup. The cup starts as the third NN and moves up to first NN in iteration 5. It is often reasonably confused with a mug. The baseline always recognized the cup as a mug in all 9 iterations, for all 3 distance metrics.

\begin{table}[thbp]
  \begin{center}
  \begin{tabulary}{.8\linewidth}{c | c c c c c c c c}
    Iter		& 1		& 2		& 3		& 4		& 5		& 6		& 7 \\ \hline
    Moves		& 14	& 14	& 17	& 18	& 7		& 5		& 2 \\
    Contacts	& 52	& 61	& 52	& 68	& 25	& 22	& 6 \\
    \hline
    1st NN		& teapot& mug	& mug	& mug	& cup	& cup	& cup \\
    1st dist	& 0.230	& 0.142	& 0.127	& 0.141	& 0.145	& 0.135	& 0.138 \\
    \hline
	2nd NN		& mug	& cup	& cup	& cup	& mug	& mug	& mug \\
	2nd dist	& 0.284	& 0.237	& 0.174	& 0.145	& 0.156	& 0.158	& 0.158 \\
    \hline
	3rd NN		& cup	&teapot	&teapot	&bottle	&bottle	&bottle	&bottle \\
	3rd dist	& 0.321	& 0.248	& 0.229	& 0.232	& 0.231	& 0.225	& 0.227
  \end{tabulary}
  \caption{\label{tab:nn} Tree result on cup}
  \end{center}
  \vspace{-5mm}
\end{table}

\subsection{Number of Moves for Recognition}


Table \ref{tab:moves} shows the number of moves per iteration for all objects. Boldface shows the iteration in which recognition starts being correct, corresponding to Fig.~\ref{fig:tree_vs_random}. Teapot was correct in iterations 1--3 and diverged to mug. All other objects stayed correct.
These are the upper bound moves for recognition, as we only ran recognition after each iteration, not after every move. Most objects were recognized within 16 moves, a large improvement over hundreds in \cite{triangles}.

\begin{table}[thbp]
  \begin{center}
  \begin{tabulary}{.8\linewidth}{c | c c c c c c c c c}
   Iteration &  1 &  2 &  3 &  4 &  5 & 6 & 7 & 8 & 9 \\ \hline
         cup & 14 & 14 & 17 & 18 &  \underline{\textbf{7}} & 5 & 2 & & \\
      teapot & \underline{\textbf{16$^*$}} & 13 & 16 & 15 &  8 & 5 & 4 & 1 & 1 \\
      bottle & \underline{\textbf{16}} & 12 & 18 & 17 &  8 & 5 & 4 & 2 & 1 \\
        bowl & \underline{\textbf{15}} & 14 & 16 & 16 &  7 & 4 & 3 & 1 & \\
         mug & \underline{\textbf{14}} & 14 & 15 & 14 & 11 & 6 & 4 & 1 & \\
toilet paper & \underline{\textbf{13}} & 18 & 15 & 15 &  8 & 5 & 1 & & \\
      sphere & 16 & 13 & 17 & 14 & 10 & 6 & 5 & 3 & 1
  \end{tabulary}
  \caption{\label{tab:moves} Upper bound number of poses to recognize correctly}
  \end{center}
  \vspace{-5mm}
\end{table}


\section{Real Robot Experiments}

\begin{figure*}[thb]
  \begin{center}
    \begin{subfigure}{.38\textwidth}
      \includegraphics[height=3.5cm]{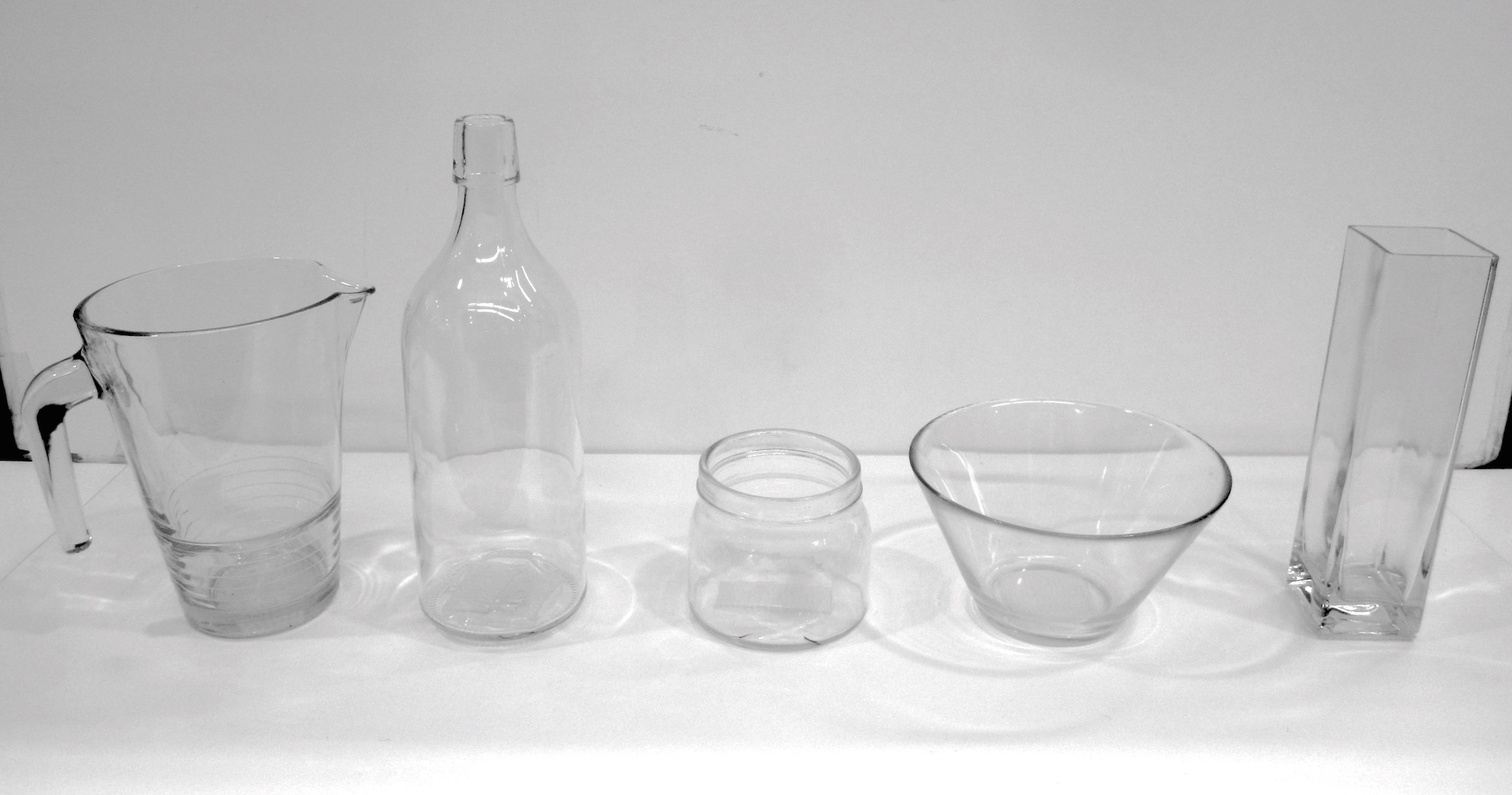}  
      \caption{}
      \label{fig:real_objs}
    \end{subfigure}
    \begin{subfigure}{.28\textwidth}
      \includegraphics[height=3.5cm,trim={0cm 0cm 0cm 1cm},clip]{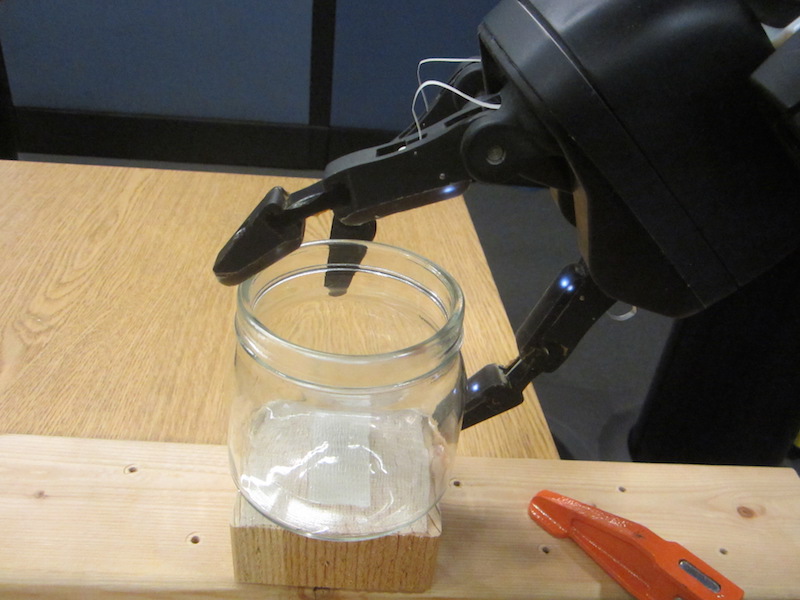}
      \caption{}
      \label{fig:real_setup}
    \end{subfigure}
    \begin{subfigure}{.32\textwidth}
      \includegraphics[width=\linewidth]{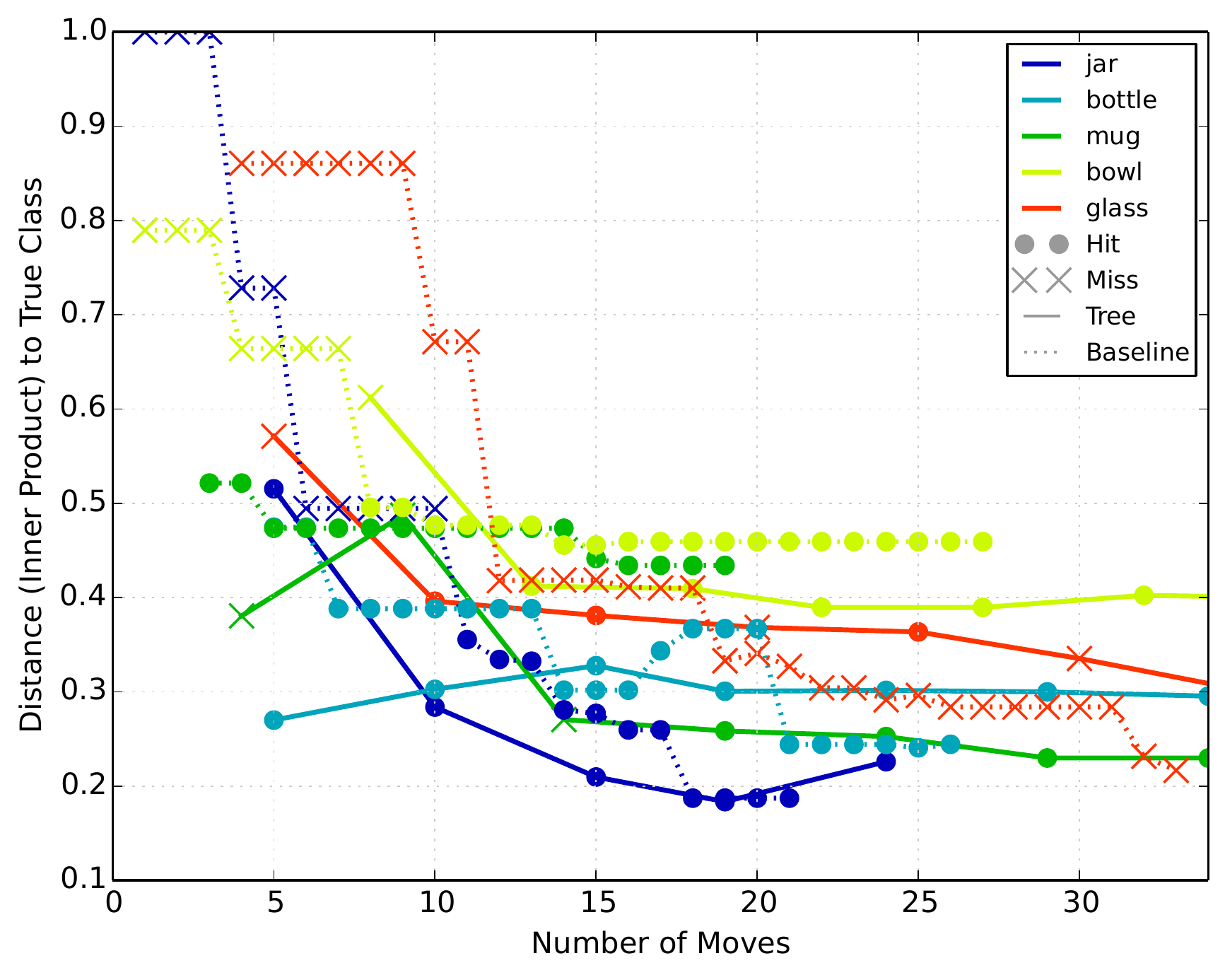}
      \vspace{-7mm}
      \caption{}
      \label{fig:real_tree_vs_greedy}
    \end{subfigure}
  \end{center}
  \vspace{-7mm}
  \caption{(a). Mug, bottle, jar, bowl, glass, used for real robot experiments. (b). Experiment setup. (c). Real robot results. Distance to true class \textit{vs.} number of poses, for baseline (dashed) and tree policy (solid). Dots and crosses show hits and misses. Each curve is an object.}
  \vspace{-5mm}
\end{figure*}

On the real robot, we compare with a greedy baseline instead of random.
Fig. \ref{fig:real_setup} shows the experiment setup.
We mounted the ReFlex Beta hand on a Baxter. 
An object is held fixed on a table. 
We trained 5 objects (Fig. \ref{fig:real_objs}) on the real robot for active instance-based recognition. Note transparent objects pose significant challenge for vision systems.

The mug, bottle, jar, bowl, and glass were each trained with 34, 60, 39, 30, and 50 end-effector poses.
With discretization (0.06 meters in translation, 0.05 in quaternion), this resulted in 138 possible actions at each MCTS node at test time.
The goal is to recognize in considerably fewer poses. This would mean the active selection is able to select poses with discriminating features.


At test time, we ran MCTS (Sec.~\ref{sec:mcts}) to actively predict a sequence of end-effector poses. Then, the Baxter arm autonomously moves to those poses, using motion planning in ROS Moveit for collision avoidance.

\begin{figure}[thbp]
  \begin{center}
  \hspace{-5mm}
  \begin{tabular}{c@{\hspace{0.2em}} c@{\hspace{0.2em}} c@{\hspace{0.2em}} c@{\hspace{0.2em}} c@{}}
    \includegraphics[height=\graspheight]{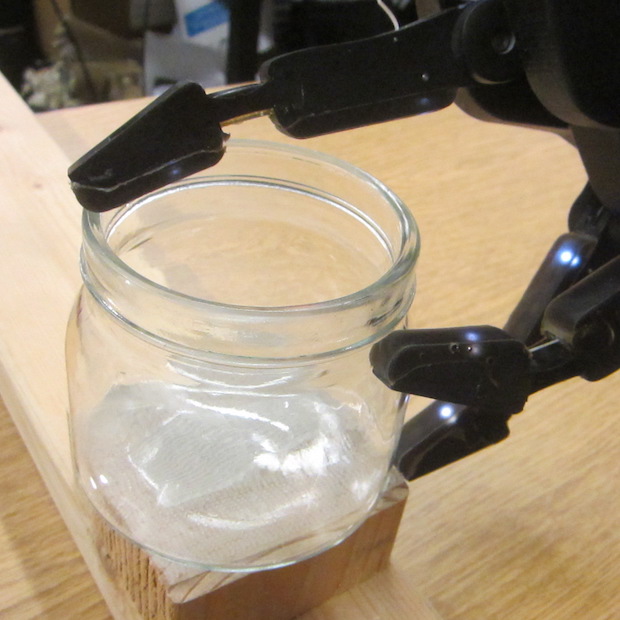} &
    \includegraphics[height=\graspheight]{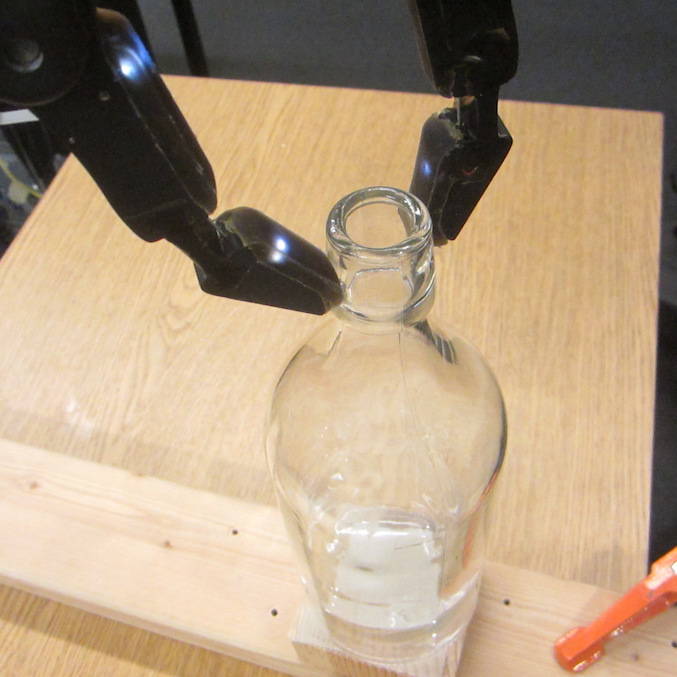} &
    \includegraphics[height=\graspheight]{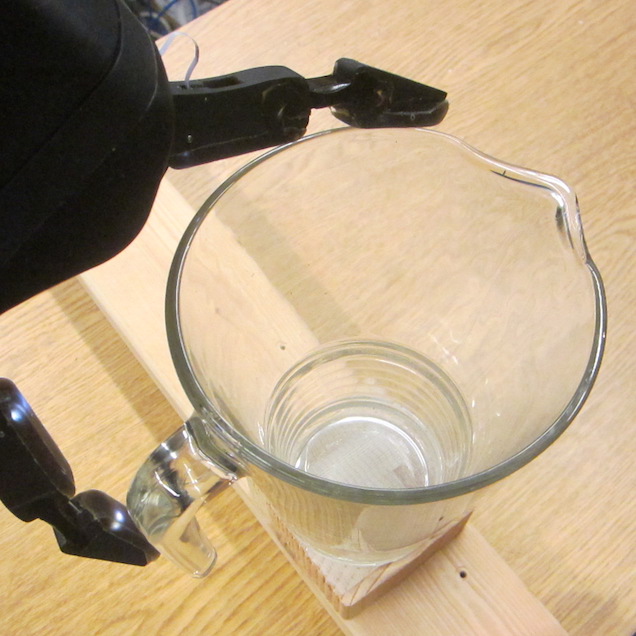} &
    \includegraphics[height=\graspheight]{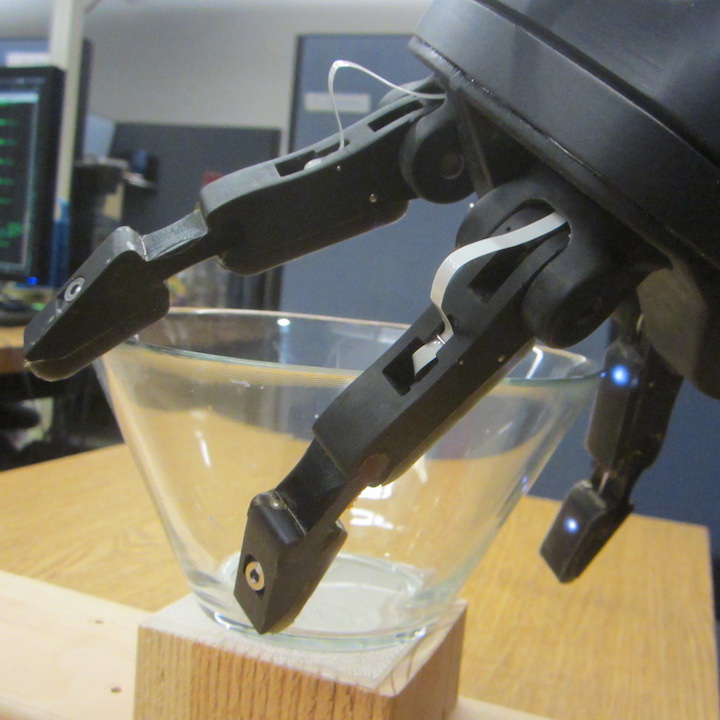} &
    \includegraphics[height=\graspheight]{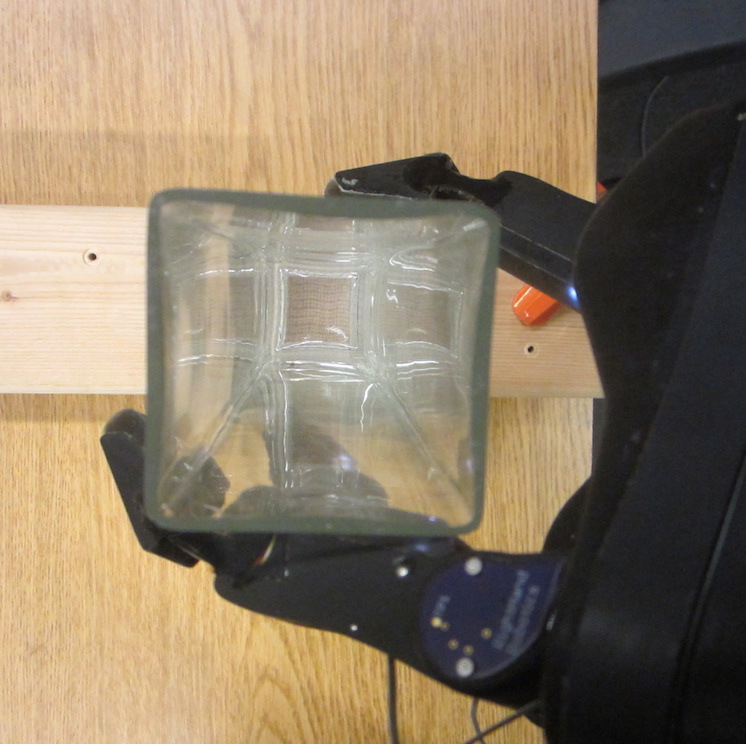}
    \\
    \includegraphics[height=\graspheight]{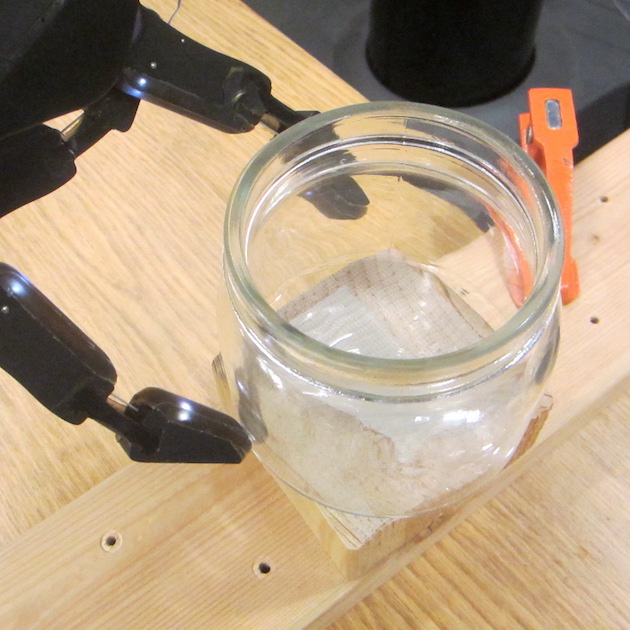} &
    \includegraphics[height=\graspheight]{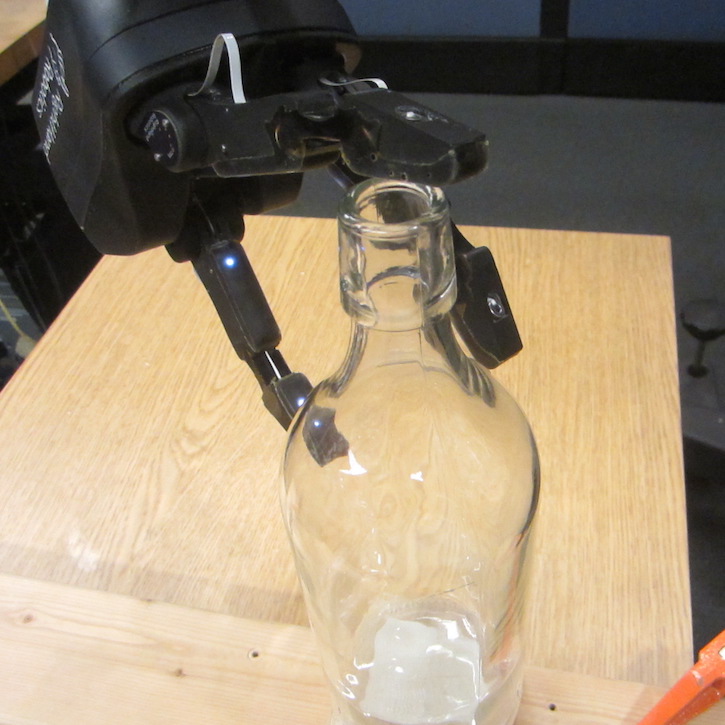} &
    \includegraphics[height=\graspheight]{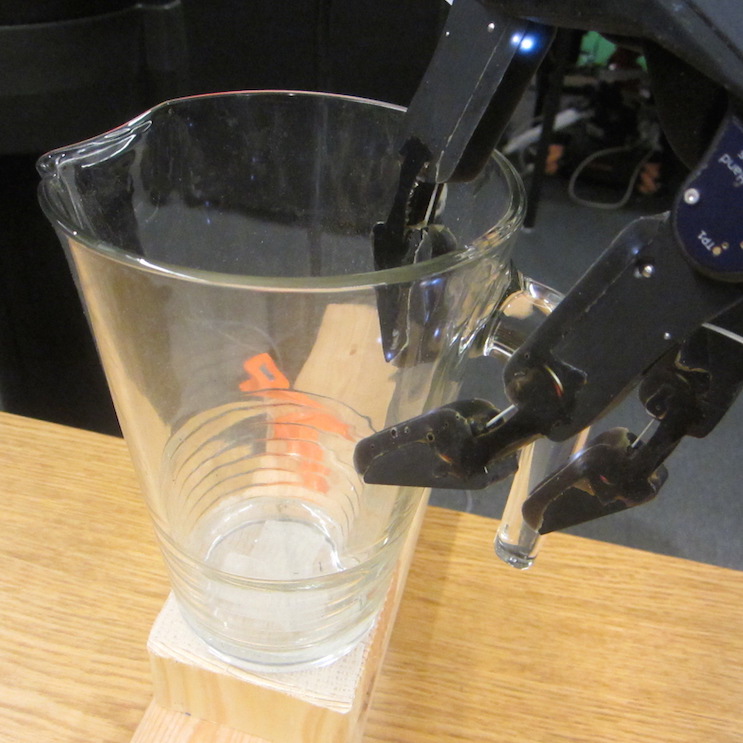} &
    \includegraphics[height=\graspheight]{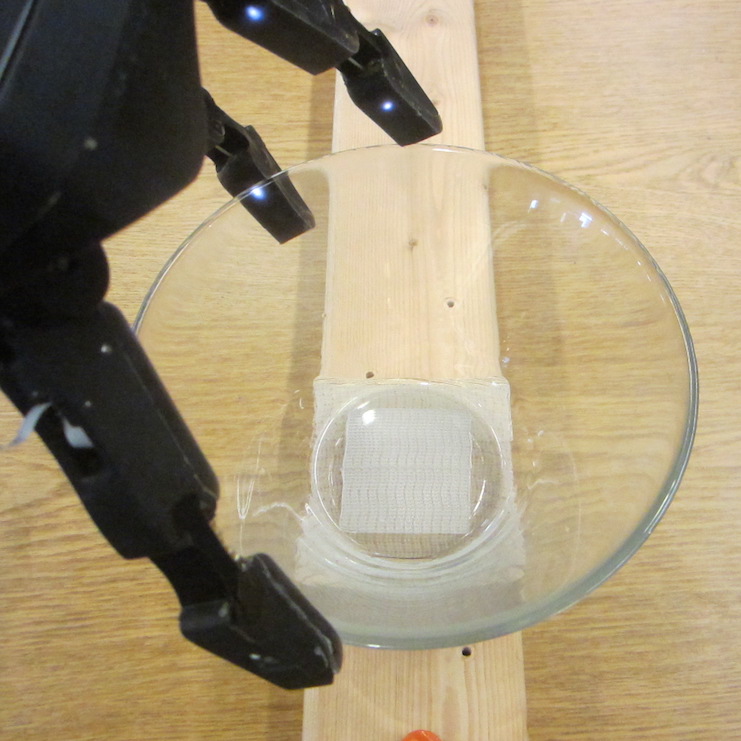} &
    \includegraphics[height=\graspheight]{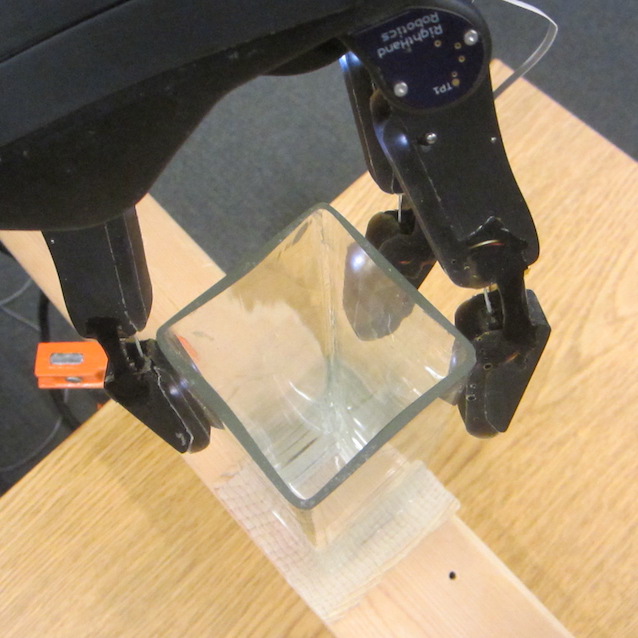}
  \end{tabular}
  \vspace{-4mm}
  \end{center}
  \caption{Real robot actions selected by tree policy at test time.}
  \label{fig:real_grasps}
\end{figure}

For a baseline, we compared with greedily selecting the immediate minimizer of the objective, \textit{i.e.} zero step lookahead, equivalent to horizon=1. For tree policy, we used horizon=5. This means tree policy had up to 5 poses per iteration; greedy had 1. We used 20 simulations per iteration for both.
Both were run until the recognition was correct for 3 consecutive iterations, some further until the distances leveled off.
Results are in Table~\ref{tab:real_tree_vs_greedy} and Fig.~\ref{fig:real_tree_vs_greedy}. Example grasps selected by tree policy are in Fig.~\ref{fig:real_grasps}. A footage with per-move distances is in the accompanying video.

Some poses selected by either method were not successfully planned, due to joint limits and collisions in the workspace.
Fig.~\ref{fig:real_tree_vs_greedy} x-axis is the raw number of poses selected.
Table~\ref{tab:real_tree_vs_greedy} shows the number of successful moves.

Tree policy recognized in significantly fewer iterations and shorter time in most cases. All objects were first recognized correctly in under 10 moves, significantly fewer than training.
Greedy never recognized the glass correctly, always as mug. Tree policy recognized it twice in a row and then flip-flopped between glass and mug.

\begin{table}[thbp]
  \begin{center}
  \begin{tabular}{c | c@{\hspace{0.8em}} c@{\hspace{0.8em}} | c@{\hspace{0.8em}} c@{\hspace{0.8em}} | c@{\hspace{0.8em}} c@{\hspace{0.8em}} c@{\hspace{0.8em}} | c@{\hspace{0.8em}} c@{\hspace{0.8em}} c@{\hspace{0.8em}}}
    Object	 & \multicolumn{4}{c}{\# Iters until Correct}  & \multicolumn{6}{c}{\# Moves until Correct} \\
     & T1 & T3 & G1 & G3 & T1 & T3 & T33 & G1 & G3 & G33 \\ \hline
    jar		& \textbf{1} & \textbf{3} & 11 & 20 &
              \textbf{2} & 10 & \textbf{10} & 
              6 & 8 & 14 \\
    bottle	& \textbf{1} & \textbf{3} & 5 & 26 &
              \textbf{2} & 8 & \textbf{8} & 
              \textbf{2} & 4 & 12 \\
    mug		& 4 & \textbf{6} & \textbf{3} & 9 &
              9 & 12 & 12 & 
              \textbf{1} & 3 & \textbf{3} \\
    bowl	& \textbf{3} & - & 9 & - &
            5 & 8 & - &
            \textbf{3} & 5 & - \\
    glass	& \textbf{2} & - & - & - &
              \textbf{7} & - & - &
              - & - & -
  \end{tabular}
  \caption{\label{tab:real_tree_vs_greedy} Number of iterations and move it took to recognize correctly on real robot. T: tree policy, G: greedy. T1/G1: first time SVM recognizing correctly; T3/G3: SVM correct 3 times in a row; T33/G33: all 3 metrics correct 3 times in a row. - denotes never.}
  \end{center}
  \vspace{-1mm}
\end{table}

\subsection{Running Time}
\label{sec:running_time}

The running time for the tree search is directly proportional to horizon $T$. Each tree simulation takes 0.5 seconds for $T=20$, 0.1 seconds for $T=5$, and 0.02 seconds for $T=1$.
Times reported are on an Intel Xeon 3.6GHz quad-core desktop simultaneously running the rest of the experiment software.
It can be improved by array access.

Even though the greedy approach ($T=1$) took shorter time per iteration, it took many more iterations before correct recognition (Table~\ref{tab:real_tree_vs_greedy}).
The reason is that $T=1$ generates only one pose per iteration, and when the pose is unfeasible due to joint limits or collision, the iteration is wasted. 
Overall, the tree policy took significantly shorter time.

\section{Conclusion}

We described an algorithm for actively selecting a sequence of end-effector poses for the objective of confident object recognition. We formulated the problem as a MDP and associated tactile observations with relative wrist poses in training, which allows the next desired actions to be predicted by observations alone at test time. The method outperforms greedily selected poses in a physics engine and on a real robot.

An improvement to optimize recognition even more directly is to select actions that would produce the most salient features in the descriptor.
Analysis methods exist for finding the most discriminative features in a classifier. The histogram descriptor makes this easy; each feature is simply a bin, which we already use as discretized observations $z$. To select the most salient action, simply select $z_{t+1}$ that maximizes saliency in addition to recognition confidence.

\bibliographystyle{IEEEtran}
\bibliography{root}

\end{document}